\documentclass[runningheads]{llncs}
\PassOptionsToPackage{table,dvipsnames}{xcolor}

\usepackage{eccv}

\usepackage{eccvabbrv}

\usepackage{graphicx}
\usepackage{booktabs,tabularx,threeparttable}
\usepackage{multirow}
\usepackage{algorithm}
\usepackage{algorithmic}
\usepackage{pdflscape} 
\usepackage{adjustbox}
\usepackage{siunitx} 
\usepackage[accsupp]{axessibility}  
\usepackage[normalem]{ulem}
\usepackage{makecell}
\usepackage{pifont} 
\usepackage{amsmath}       

\usepackage[hidelinks]{hyperref}

\usepackage{orcidlink}

\newcommand{\cmark}{\ding{51}} 
\newcommand{\xmark}{\ding{55}} 

\begin{document}

\title{Anomaly Factory 3D: A Modular Framework for Diverse Pseudo-Anomaly Synthesis in Unsupervised 3D Anomaly Detection} 

\titlerunning{AF3AD}

\author{Ali Balapour\inst{1,2,3}\orcidlink{0000-0002-4552-2094} \and
Faraz Hach\inst{1,2,3}\orcidlink{0000-0003-1143-0172}\thanks{Corresponding author: \texttt{faraz.hach@ubc.ca}.}}

\authorrunning{A.~Balapour and F.~Hach}

\institute{
University of British Columbia, Vancouver, V6T 1Z4 BC, Canada \and
Vancouver Prostate Centre, M.H. Mohseni Institute of Urologic Sciences, Vancouver, V6H 3ZB BC, Canada \and 
Department of Urologic Sciences, The University of British Columbia, Vancouver, V5Z 1M9 BC, Canada
}

\maketitle

\begin{abstract}
Detecting and localizing defects in 3D point clouds is challenging because abnormal samples are scarce and diverse, while training is often limited to normal data. We propose Anomaly Factory 3D (AF3AD), a modular framework that synthesizes diverse pseudo-anomalies from normal point clouds to expand the training data for unsupervised 3D anomaly detection methods that rely on pseudo-anomalies. AF3AD uses a center-conditioned parametric deformation model defined in local PCA frames, with kernel-controlled spatial falloff, anisotropy, directional gating, and normal/tangential displacement fields, enabling a broad set of geometric defect presets. We demonstrate its ease-of-use and effectiveness by integrating AF3AD with an offset-prediction detector and a reconstruction-based anomaly detection method, showing that AF3AD transfers across detection paradigms. Experiments on AnomalyShapeNet and Real3D-AD show consistent improvements in object- and point-level detection and localization, supported by ablations on preset groups and robustness under noise. AF3AD is designed as a standalone synthesis tool to facilitate adoption across different 3D anomaly detection paradigms. Code is available at github.com/vpc-ccg/AF3AD.
\end{abstract}

\section{Introduction}
\label{sec:intro}

Anomaly detection is a critical task across multiple domains, including natural language, audio, and image data~\cite{DL_Advancements_in_AD}. However, the 3D  shape modality remains less explored, despite being essential for detailed analysis in scientific and engineering applications. This gap exists largely because 3D data are difficult to capture and computationally intensive to process. Furthermore, 3D anomaly detection (3DAD) is fundamentally more challenging than its 1D (time-series) or 2D (images) counterparts due to structural complexities such as geometric sparsity, viewpoint dependence, and sensor noise~\cite{3DAD_manufacturing_survey,cao2024surveyvisualanomalydetection}.

Despite these technical challenges, performing anomaly detection on 3D data offers distinct advantages over traditional 2D images.
These include a richer understanding of an object’s spatial geometry, reduced reliance on a single viewpoint, and lower sensitivity to the illumination variations that frequently affect 2D data~\cite{3DAD_manufacturing_survey}.
One primary use case for 3DAD is found in industrial settings~\cite{MVTec3D}, particularly for identifying defects in manufactured objects.

The core task of 3DAD involves identifying regions on the surface of a 3D object, particularly within point cloud representations, that deviate from normal patterns.
These deviations can range from simple surface bulges to complex structural deformations.
To identify these areas, Artificial Intelligence (AI) based methods have proven particularly effective for both detection and localization.
Recent advancements in these methods have enabled faster and more accurate detection, with some approaches now achieving real-time defect detection on 3D data~\cite{uiad_survey,MiniShift}.
AI-based methods for 3DAD can generally be grouped into three main paradigms: (1) reconstruction-based approaches, which learn a model of normal geometry and flag deviations via reconstruction residuals; (2) memory-bank (feature-based) methods, which store representations of normal samples and detect anomalies as feature mismatches at inference time; and (3) multi-view approaches, which render 3D data into a set of 2D views to leverage mature 2D anomaly detection pipelines.

Due to the scarcity of anomalous data, training 3DAD methods directly on defective samples is often impractical.
As a result, many works adopt an unsupervised setting, where only normal samples are available during training and the model is not exposed to anomalous observations.
This paradigm is commonly known as Unsupervised 3D Anomaly Detection (U3DAD) \cite{cao2024surveyvisualanomalydetection}.
Within this paradigm, an increasingly adopted strategy is to synthesize pseudo-anomalies from normal data to create surrogate supervision.
Recent methods such as R3D-AD~\cite{R3DAD} and PO3AD~\cite{PO3AD} follow this direction and demonstrate that learning to distinguish normal patterns from generated defects can improve detection at inference time.
However, existing synthesis pipelines are often limited in the diversity and structural complexity of the generated defects, which may restrict the range of abnormal cues the model learns.
This motivates a more structured pseudo-anomaly generation framework that can produce richer and more realistic defect patterns to better support robust 3D anomaly detection.

We propose \textbf{A}nomaly \textbf{F}actory for \textbf{3}D \textbf{A}nomaly \textbf{D}etection (\textbf{AF3AD}), a method designed to synthesize a diverse set of pseudo-anomalies using geometric kernels.
These kernels are controlled by probabilistic distributions to introduce greater variation into the synthesis process.
Using parameterized kernels, we design a repertoire of pseudo-anomalies ranging from simple bulges and dents to directional drags and elongated ridges.
These are applied to normal, non-anomalous 3D point cloud samples to prepare them for unsupervised anomaly detection training.
Our framework is a standalone, reusable module that supports a wide range of parameter settings, such as radius and displacement rate, enabling greater diversity in the generated anomalies.
To validate the effectiveness of our synthesis framework, we instantiate and evaluate it with an offset-prediction detector based on PO3AD~\cite{PO3AD} and a reconstruction-based detector based on R3D-AD~\cite{R3DAD}. 
We show that using more diverse and complex anomalies increases the accuracy and precision on established benchmarks.
We evaluate our method on two prominent datasets, AnomalyShapeNet~\cite{AnomalyShapeNet} and Real3D-AD~\cite{Real3DAD}, demonstrating that our integration achieves competitive performance.
Furthermore, we discuss the advantages that synthetic diversity provides for the task of unsupervised anomaly detection.

In summary, our main contributions are as follows: (1) We propose AF3AD, a modular synthesis framework that generates diverse geometric pseudo-anomalies with explicit parametric control, designed to enhance 3D anomaly detection methods that rely on pseudo-anomaly synthesis. (2) When integrated with offset prediction detector, we achieve competitive performance, reaching 91.5\% O-AUROC on AnomalyShapeNet (+5.5 points) and 85.2\% on Real3D-AD (+7.1 points). (3) We further provide extensive ablations demonstrating that increased synthesis diversity consistently improves detection performance, remains robust under noise, and is supported by an empirical feature-space alignment analysis.
\section{Related Works}

\subsection{2D Anomaly Detection}
Unsupervised 2D visual anomaly detection, where only normal samples are available during training, has advanced considerably in recent years and can be grouped into four main paradigms~\cite{cao2024surveyvisualanomalydetection}.
Memory-bank methods store feature representations of normal data and detect anomalies by measuring their deviation from learned patterns~\cite{patchcore}.
Reconstruction-based approaches identify anomalies through errors in reconstructing normal samples or features~\cite{DFR}.
Knowledge-distillation methods use teacher--student discrepancies as anomaly signals~\cite{adrd}, while flow-based models detect anomalies via low likelihood under an explicitly learned distribution of normal features~\cite{cflow-ad}.
Despite their success in 2D image-based settings, directly applying these paradigms to 3D data remains challenging because point clouds are unstructured, unordered, and sparse, unlike the regular grids of images.

\subsection{3D Anomaly Detection}

3D anomaly detection aims to detect and localize geometric irregularities (e.g., structural deformations or surface defects) in unstructured point clouds~\cite{MVTec3D}.
Earlier methods relied on handcrafted descriptors like FPFH~\cite{BTF} to capture local geometry invariant to rotation.
With the advent of deep learning, teacher–student architectures (e.g., TS-AD~\cite{tsad}) emerged, distilling features from pre-trained encoders like PointNet.
Recent approaches leverage large-scale pre-trained transformers; for instance, PointMAE~\cite{PointMAE} and PointBERT~\cite{pointbert} are frequently used as backbones to extract robust semantic features from raw geometry.
Building on these representations, recent works have diversified: Reg3D-AD~\cite{Real3DAD} utilizes registration to align test samples with a normal template, while Reg2Inv~\cite{Reg2Inv} explicitly learns registration-invariant representations to handle pose variations.
Other strategies include bridging the modality gap by fusing 2D multi-view images with 3D descriptors (CPMF~\cite{CPMF}), or explicitly modeling anomaly magnitude via per-point offset regression using sparse backbones (PO3AD~\cite{PO3AD}).

\subsection{Pseudo-Anomaly Synthesis}  

To mitigate the scarcity of anomalous training data, a body of work synthesizes pseudo-anomalies directly on 3D geometry.
While earlier multimodal methods, such as CLIP3D-AD~\cite{clip3d-ad}, applied 2D noise, geometric methods focus on structural manipulation.
Patch-Gen in R3D-AD~\cite{R3DAD} introduces patch-based rotation and translation, as well as bulges, dents, and random damage, while GLFM~\cite{glfm} stretches points along normal vectors to simulate protrusions.
Addressing unrealistic overlaps, Norm-AS in PO3AD~\cite{PO3AD} constrains displacements such as bulges and dents along surface normals.
Despite these advances, current synthesis methods often lack the structural diversity required for robust generalization, motivating the need for richer generation strategies (see the summary in \Cref{tab:synthesis_comparison}). 
MiniShift~\cite{MiniShift} also uses geometric synthesis, but for building a fixed benchmark rather than generating training supervision. Therefore, it is complementary to AF3AD.

\begin{table}[t]
\centering
\caption{Comparison of pseudo-anomaly synthesis methods for 3D anomaly detection.}
\label{tab:synthesis_comparison}
\small
\begin{tabular}{lcccc}
\toprule
Method & Deformation & Anisotropic & Directional & Modular \\
       & Diversity   & Support     & Control     & Design \\
\midrule
CLIP3D-AD & Low (1 type) & \xmark & Normal only & \xmark \\
Patch-Gen (R3D-AD) & Low (3 types) & \xmark & Random & \xmark \\
GLFM & Low (1 type) & \xmark & Normal only & \xmark \\
Norm-AS (PO3AD) & Low (2 types) & \xmark & Normal only & \xmark \\
\midrule
\textbf{AF3AD (Ours)} & \textbf{High (11 types)} & \cmark & \textbf{Normal+Tangential} & \cmark \\
\bottomrule
\end{tabular}
\vspace{-0.1in}
\end{table}

\section{Methods}

\subsection{Overview}
Our framework is modular and operates independently of specific detection architectures.
AF3AD is designed to address a key limitation of prior pseudo-anomaly synthesis methods: tight coupling with specific detection paradigms.
For instance, Norm-AS~\cite{PO3AD} is designed for offset regression, whereas Patch-Gen~\cite{R3DAD} is tailored to diffusion-based reconstruction.
Such coupling reduces reusability and limits systematic study of synthesis strategies.
In contrast, AF3AD acts as a standalone synthesis module that transforms normal point clouds into diverse pseudo-anomalous samples with corresponding anomaly masks.
These synthetic samples can be used to supervise detectors trained with pseudo-anomaly injection, including reconstruction-, registration-, and offset-based methods.
We instantiate AF3AD with both offset-prediction (\Cref{subsection:offset_pred}) and reconstruction-based (\Cref{sec:instantiation_r3dad}) detectors, although the synthesis module itself (\Cref{subsection:pseudo_anoamly_synthesis}) is designed to remain standalone.

\begin{figure}[t!]
    \centering

    \includegraphics[width=\linewidth]{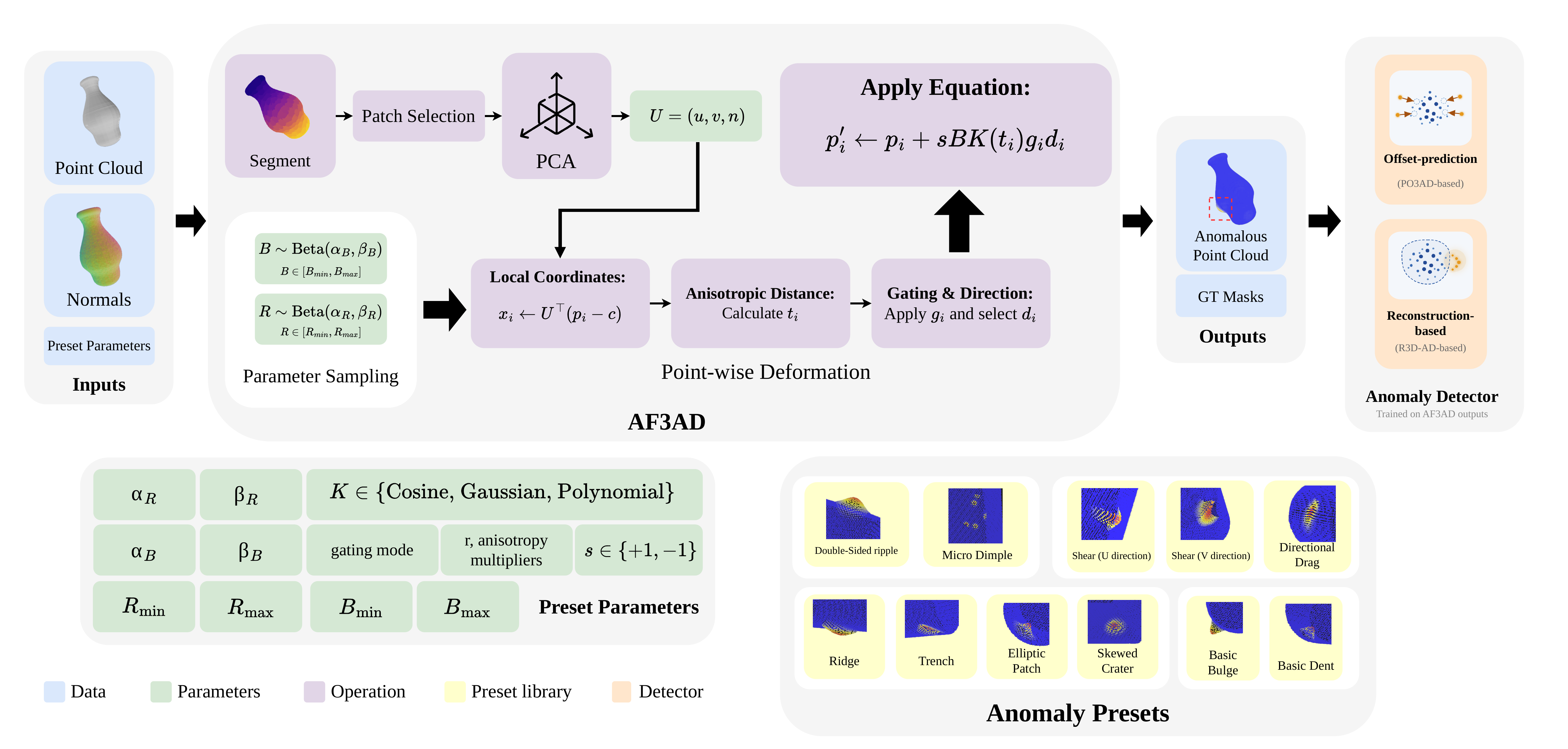}
    
    \caption{Overview of AF3AD pipeline for geometric pseudo-anomaly synthesis. From a non-anomalous point cloud and its surface normals, AF3AD selects a local segment and computes a PCA-based local frame. Within this segment, each point is transformed into local coordinates, evaluated through an anisotropic distance function, and displaced according to preset-specified gating, direction, and deformation parameters. This modular design enables the controlled synthesis of diverse geometric pseudo-anomalies, producing an anomalous point cloud along with preset-specific anomaly patterns and ground-truth (GT) masks.}
    \label{fig:architecture}
\end{figure}

\subsection{Pseudo-Anomaly Synthesis via Parametric Deformations}
\label{subsection:pseudo_anoamly_synthesis}

Pseudo-anomalies are synthesized by applying localized, parametric deformations to a non-anomalous 3D point cloud. 
Let $P=\{\mathbf{p}_i\}_{i=1}^{M}\subset\mathbb{R}^3$ denote a point cloud with $M$ points and 
$N=\{\mathbf{\hat{n}}_i\}_{i=1}^{M}$ the corresponding surface normals. 
The point cloud is partitioned into $K = 64$ patches by randomly sampling $K$ centers and assigning each point to its nearest center (single pass, no $k$-means); a random patch center $c \in R^3$ then defines the deformation region.
This produces a deformed point cloud $P'=\{\mathbf{p}'_i\}_{i=1}^{M}$ by applying a displacement field to each point in that segment.

\subsubsection{General Deformation Formulation} In the generated pseudo-anomalous sample $P'$, each point is displaced as
\begin{equation}
\mathbf{p}'_i 
= 
\mathbf{p}_i 
+ 
\Delta(\mathbf{p}_i), 
\qquad 
\Delta(\mathbf{p}_i) 
= 
s \, B \, w_i \, \mathbf{d}_i,
\end{equation}
where $s\in\{+1,-1\}$ controls the deformation sign (e.g., inward vs. outward), 
$B>0$ controls the maximum displacement magnitude, 
$w_i\in[0,1]$ is a spatial falloff weight, and 
$\mathbf{d}_i\in\mathbb{R}^3$ denotes the unit deformation direction vector. 
This parameterized formulation provides control and flexibility to build a diverse set of pseudo-anomalies.
Points outside the selected segment remain unchanged, i.e., $\mathbf{p}'_i=\mathbf{p}_i$ for points not assigned to the deformed segment.

\subsubsection{Local Coordinate Frame and Anisotropic Support} 
To define anisotropic deformations, we estimate a local orthonormal frame at 
the anomaly center $\mathbf{c}$ using local Principal Component Analysis (PCA)~\cite{pca}:
\begin{equation}
U(\mathbf{c}) = [\mathbf{u}\;\mathbf{v}\;\mathbf{n}] \in \mathbb{R}^{3\times 3},
\qquad
U^\top U = I,
\end{equation}
where $\mathbf{u}$ and $\mathbf{v}$ span the principal tangent directions of the local neighborhood, 
and $\mathbf{n}$ is the local mean normal. 
Points are then expressed in this local frame as
\begin{equation}
\mathbf{x}_i = U^\top(\mathbf{p}_i - \mathbf{c}).
\end{equation}

Let $R$ denote a nominal anomaly radius, and let $(r_u, r_v, r_n)$ be dimensionless anisotropy multipliers. 
We define an ellipsoidal normalized distance
\begin{equation}
t_i 
= 
\left(
\frac{x_{i,u}^2}{(r_u R)^2} 
+ 
\frac{x_{i,v}^2}{(r_v R)^2} 
+ 
\frac{x_{i,n}^2}{(r_n R)^2}
\right)^{1/2},
\end{equation}
which induces anisotropic spatial support around the anomaly center.

\subsubsection{Spatial Falloff Kernels} 
Each point is assigned a spatial weight through a kernel function
\begin{equation}
w_i = K(t_i), 
\qquad 
w_i \in [0,1].
\end{equation}
We employ smooth kernels such as cosine, Gaussian, or polynomial functions, optionally parameterized 
by shape parameters (e.g., $q$ or $\sigma$), to ensure spatially coherent deformations and avoid 
isolated point displacements.

To support asymmetric deformations, we optionally apply a soft gating function $g_i\in[0,1]$ and set
$w_i \leftarrow w_i g_i$, where $g_i$ is either a half-space gate in a chosen direction or
a normal-alignment gate restricting deformation to similarly oriented surfaces.
We choose $\mathbf{d}_i \in \{\mathbf{n},\mathbf{u},\mathbf{v}\}$ (from principal directions).
When gating is disabled, we set \(g_i=1\) for all points.

\subsubsection{Stochastic Parameter Sampling} 
The sign parameter $s \in \{+1, -1\}$ determines the direction of the deformation (e.g., $+1$ for an outward bulge and $-1$ for an inward dent).
This parameter is either fixed for a specific anomaly preset or sampled stochastically during synthesis.
Both the deformation magnitude $B$ and the nominal anomaly radius $R$ are sampled 
from bounded Beta distributions to introduce controlled stochasticity while preserving geometric 
plausibility. 
Specifically, for a parameter $z \in \{B, R\}$, we sample
\begin{equation}
z 
= 
z_{\min} 
+ 
(z_{\max} - z_{\min}) 
\cdot 
\mathrm{Beta}(\alpha_z, \beta_z),
\end{equation}
where $\mathrm{Beta}(\alpha_z, \beta_z)$ denotes the Beta distribution with shape parameters 
$(\alpha_z, \beta_z)$ and support on $[0,1]$. 
The bounds $(z_{\min}, z_{\max})$ and distribution parameters $(\alpha_z, \beta_z)$ are specified 
per anomaly preset, allowing precise control over the typical scale and severity of synthesized 
defects while preventing extreme or implausible deformations.

Combining all components, each point is transformed as
\begin{equation}
\boxed{
\mathbf{p}'_i 
= 
\mathbf{p}_i 
+ 
s \, B \, K(t_i) \, g_i \, \mathbf{d}_i
.}
\end{equation}

\subsubsection{Preset Taxonomy}
This formulation unifies isotropic and anisotropic bulges and dents, elongated ridges and trenches, skewed impact craters, tangential shear and drag anomalies, and compound micro-defect fields under a single-center-conditioned deformation model.
A summary of the defined presets is provided in \Cref{tab:anomaly_taxonomy}, and the overall procedure is outlined in \Cref{fig:architecture}.
We have grouped these anomalies based on their features (e.g., direction, anisotropy, etc.) for further analysis.
Additionally, AF3AD is not limited to these specific presets; additional deformation types can be easily incorporated into the framework.

\begin{table}[t]
\centering
\begin{threeparttable}
\caption{Pseudo-anomaly preset taxonomy.}
\label{tab:anomaly_taxonomy}

\renewcommand{\arraystretch}{1.1} 

\begin{tabular}{@{} l c c @{}}
\toprule
\textbf{Anomaly Type} & \textbf{Aniso.} & \textbf{Sym.} \\
\midrule
\multicolumn{3}{@{}l}{\textbf{Group A} --- \textit{Normal-direction, isotropic single-lobe}} \\
\hspace{1em} Basic Bulge & Iso & 1-side \\
\hspace{1em} Basic Dent  & Iso & 1-side \\
\midrule 
\multicolumn{3}{@{}l}{\textbf{Group B} --- \textit{Normal-direction, anisotropic structured marks}} \\
\hspace{1em} Elongated Ridge      & Aniso & 1-side \\
\hspace{1em} Trench / Groove      & Aniso & 1-side \\
\hspace{1em} Elliptic Flat Spot   & Mild  & 1-side \\
\hspace{1em} Skewed Impact Crater & Aniso & Dir.   \\
\midrule
\multicolumn{3}{@{}l}{\textbf{Group C} --- \textit{Normal-direction, multi-frequency/periodic}} \\
\hspace{1em} Double-Sided Ripple  & Iso & 2-side \\
\hspace{1em} Micro-Dimple Field   & Iso & 1-side \\
\midrule
\multicolumn{3}{@{}l}{\textbf{Group D} --- \textit{Tangential-direction surface shifts}} \\
\hspace{1em} Shear / Slip ($\mathbf{u}$)    & Iso  & 2-side \\
\hspace{1em} Shear / Slip ($\mathbf{v}$)    & Iso  & 2-side \\
\hspace{1em} Directional Drag / Stretch     & Aniso & 1-side \\
\bottomrule
\end{tabular}

\begin{tablenotes}[flushleft]\footnotesize
\item \textit{Note:} Aniso.\ = anisotropy (Iso: isotropic, Aniso: anisotropic, Mild: mildly anisotropic); Sym.\ = symmetry (1-side: one-sided, 2-side: two-sided, Dir.: direction-gated).
\end{tablenotes}

\end{threeparttable}
\end{table}


\subsection{Instantiation: Offset-Prediction Detector}
\label{subsection:offset_pred}
To validate the effectiveness of our modular synthesis framework, we integrate it with an offset-prediction detection architecture.
We build upon PO3AD~\cite{PO3AD} with architectural refinements to better leverage our diverse pseudo-anomalies, replacing the Norm-AS module proposed in that work.

After synthesizing pseudo-anomalies on training samples using AF3AD, we train a modified version of the PO3AD model.
In this approach, the model predicts an offset for every point in the point cloud, representing its displacement from its expected normal position (i.e., $O_x, O_y, O_z$).
An offset of $O_x=0, O_y=0, O_z=0$ indicates that a point is in its original, non-anomalous position, while any non-zero value suggests displacement in 3D space by the corresponding amount.
\subsubsection{Anomaly Score Calculation.} For each point in the point cloud, we compute an individual anomaly score using the $L_1$ norm of its predicted offset (i.e., $|O_x| + |O_y| + |O_z|$).
This provides a straightforward measure of point-wise deviation, where a higher score indicates a greater likelihood of an anomaly at that specific location.
To determine the overall anomaly score for the entire point cloud (i.e., for object-level detection), we simply calculate the sum of these individual point-level scores.
Detailed training procedures are provided in the supplementary material (\Cref{supp:offset_detector}).

\subsection{Instantiation: Reconstruction-based Anomaly Detection}
\label{sec:instantiation_r3dad}

To demonstrate AF3AD’s modularity and ease of integration, we instantiate it within R3D-AD~\cite{R3DAD} as a second reference framework for anomaly detection.
R3D-AD reconstructs the geometry of 3D point clouds using an autoencoder architecture with a PointNet-based encoder and a diffusion-based decoder.
During training, it leverages pseudo-anomalous samples generated from normal data via a module called Patch-Gen.
Patch-Gen synthesizes three anomaly types: bulges, dents, and damages (random point perturbations), by selecting a random view of the point cloud and extracting local segments using a $k$-nearest-neighbors ($k$NN) search.
In contrast, AF3AD is geometry-driven and view-independent, operating directly on the 3D structure of the point cloud.
In this instantiation, we replace Patch-Gen with AF3AD to generate pseudo-anomalous training samples while keeping the remaining R3D-AD architecture and training pipeline unchanged.

\section{Experiments}
\subsection{Experimental Settings}

\subsubsection{Datasets}
We evaluated our method on two standard benchmark datasets: AnomalyShapeNet~\cite{AnomalyShapeNet} and Real3D-AD~\cite{Real3DAD}.
AnomalyShapeNet, derived from ShapeNet~\cite{ShapeNet}, contains 40 categories and 1,600 samples with synthetically introduced anomalies.
The training set includes only four normal samples per category, while the test set contains both normal and anomalous samples. Real3D-AD contains 12 categories of high-resolution point clouds scanned from real objects.
Its training set includes four normal samples per category, and its test set consists of 100 normal and anomalous samples per category.
A key difference between the datasets is scan coverage: Real3D-AD uses full 360-degree scans for training but single-view scans for testing, whereas AnomalyShapeNet uses full scans for both.
This train--test discrepancy in Real3D-AD introduces a domain shift that makes evaluation more challenging.

\subsubsection{Evaluation Metrics}
To evaluate 3D anomaly detection performance, it is common practice to use the Area Under the Receiver Operating Characteristic Curve (AUROC).
We utilize object-level AUROC (O-AUROC) to assess anomaly detection for the entire object, and point-level AUROC (P-AUROC) to evaluate anomaly localization on the point cloud, independent of a specific decision threshold.
Furthermore, to provide a more comprehensive analysis, we report the Area Under the Precision-Recall Curve (AUPR) at both levels (O-AUPR and P-AUPR) whenever possible.
Finally, when comparing with existing methods in the literature, we provide the average rank as an additional metric to facilitate detailed multi-method comparisons.

\subsection{Implementation Details}
The anomaly detector uses a MinkUNet34C~\cite{mink} backbone and a three-head offset MLP with hidden size 128 and PReLU activations. We matched training exposure across datasets to 150{,}000 effective epochs and trained using AdamW at an initial learning rate of $5 \times 10^{-4}$. For Real3D-AD, we used cosine annealing with 50 warmup epochs and a two-stage downsampling strategy (voxel grid sampling followed by farthest point sampling (FPS)), which improved efficiency and occasionally improved performance (\Cref{tab:real3dad_downsample_ablation_summary,subsubsection:ablation_downsampling}). Input point clouds were centered, normalized, randomly rotated, and processed using 64 patches, with normal and tangent directions extracted from the provided \texttt{.obj} and \texttt{.ply} files. Profiling on AnomalyShapeNet shows that AF3AD is lightweight, requiring 0.96 ms per call, comparable to Norm-AS (1.02 ms) and faster than Patch-Gen (3.24 ms), with negligible impact on training throughput.

\subsection{Baseline Methods}
We compare our method with multiple prominent works in the literature: BTF~\cite{BTF} (using raw and FPFH features), M3DM~\cite{M3DM}, PatchCore~\cite{patchcore} (using FPFH features or the PointMAE~\cite{PointMAE} backbone), 
ISMP~\cite{Liang2024LookIF}, CPMF~\cite{CPMF}, Reg3D-AD~\cite{Real3DAD}, IMRNet~\cite{AnomalyShapeNet}, R3D-AD~\cite{R3DAD}, 
Group3AD~\cite{Group3AD}, PO3AD~\cite{PO3AD}, and Reg2Inv~\cite{Reg2Inv}.
Note that we only include Group3AD on Real3D-AD. The results for these methods were obtained either from the original papers or from publicly available implementations.

\subsection{Main Results}
\label{sec:results}

\subsubsection{Results on AnomalyShapeNet}
\Cref{tab:combined_summary_auroc_rank} summarizes the results of our proposed framework on the AnomalyShapeNet dataset in comparison to baseline methods.
We report the performance of AF3AD as the mean and standard deviation over three random seeds.
The average rank for AF3AD is computed using the per-category mean over three seeds.
AF3AD achieves the best performance overall at both object and point levels, with 91.5\% O-AUROC and 92.5\% P-AUROC.
It outperforms the best competing methods by 5.5 points in O-AUROC (vs. Reg2Inv at 86.0\%) and 2.6 points in P-AUROC (vs. PO3AD at 89.9\%). Compared to PO3AD, AF3AD improves O-AUROC by 7.6 points and P-AUROC by 2.6 points. Detailed class-wise results are provided in Supplementary \Cref{tab:full_ANS_oauroc_results,tab:full_ANS_pauroc_results}.

\begin{table}[t] \centering \footnotesize \setlength{\tabcolsep}{4pt} \caption{Summary of mean AUROC and average rank on AnomalyShapeNet and Real3D-AD. AF3AD is reported as mean $\pm$ std (\%) over three different seeds. Higher AUROC is better ($\uparrow$); lower rank is better ($\downarrow$). The results of the offset-prediction instantiation are reported. Best is \textbf{bold}, second-best is \uline{underlined}.} \label{tab:combined_summary_auroc_rank} \begin{adjustbox}{width=\linewidth} \begin{tabular}{@{}l cccc cccc @{}} \toprule & \multicolumn{4}{c}{AnomalyShapeNet} & \multicolumn{4}{c}{Real3D-AD} \\ \cmidrule(lr){2-5} \cmidrule(lr){6-9} Method & O-AUROC $\uparrow$ & P-AUROC $\uparrow$ & O-Rank $\downarrow$ & P-Rank $\downarrow$ & O-AUROC $\uparrow$ & P-AUROC $\uparrow$ & O-Rank $\downarrow$ & P-Rank $\downarrow$ \\ \midrule BTF(RAW) & 49.3 & 55.0 & 9.70 & 9.76 & 63.5 & 57.1 & 9.21 & 9.42 \\ BTF(FPFH) & 52.8 & 62.8 & 9.00 & 7.55 & 60.3 & 73.3 & 9.25 & 6.33 \\ M3DM & 55.2 & 61.6 & 8.78 & 7.88 & 55.2 & 63.1 & 11.62 & 8.75 \\ PatchCore(FPFH) & 56.8 & 58.0 & 8.34 & 8.78 & 59.3 & 68.2 & 10.00 & 6.92 \\ PatchCore(PMAE) & 56.2 & 57.7 & 8.39 & 9.00 & 59.4 & 62.0 & 10.58 & 9.00 \\ CPMF & 55.9 & 57.3 & 8.31 & 9.35 & 58.6 & 75.9 & 10.58 & 5.29 \\ Reg3D-AD & 57.2 & 66.8 & 8.43 & 6.25 & 70.4 & 70.5 & 7.50 & 6.83 \\ Group3AD & -- & -- & -- & -- & 75.1 & 73.5 & 5.83 & 5.17 \\ IMRNet & 65.9 & 64.9 & 5.88 & 6.86 & 72.5 & -- & 6.58 & -- \\ ISMP & -- & 69.1 & -- & 6.05 & 75.7 & 83.6 & 5.42 & 3.21 \\ R3D-AD & 74.9 & -- & 4.06 & -- & 73.4 & -- & 6.17 & -- \\ PO3AD & 83.9 & \uline{89.9} & 2.94 & \uline{2.35} & 76.6 & -- & 5.33 & -- \\ Reg2Inv & \uline{86.0} & 88.2 & \uline{2.56} & 2.52 & \uline{78.1} & \textbf{87.9} & \uline{4.42} & \uline{2.58} \\ \midrule PO3AD+AF3AD (Ours) & \textbf{91.5$\pm$2.1} & \textbf{92.5$\pm$1.6} & \textbf{1.62} & \textbf{1.65} & \textbf{85.2$\pm$1.2} & \uline{86.1$\pm$1.9} & \textbf{2.50} & \textbf{2.50} \\ \bottomrule \end{tabular} \end{adjustbox} \end{table}

\subsubsection{Results on Real3D-AD}
\Cref{tab:combined_summary_auroc_rank} summarizes the results of our proposed framework on the Real3D-AD dataset compared to baseline methods.
To address the train--test discrepancy in scan coverage (full scans during training versus partial scans during testing), we additionally investigated different augmentation settings during preprocessing to mimic partial scans in training data, with details provided in \Cref{app:real3dad_train_data_ablation,tab:real3dad_train_data_ablation}.
Our method outperforms the second-best approach in O-AUROC by a margin of 7.1 points and achieves competitive results at the point level (second in raw P-AUROC, best in average point-level rank).
Notably, our framework achieves an 8.6 O-AUROC points gain over PO3AD, which serves as the foundational method for our work.
Detailed class-wise results are provided in Supplementary \Cref{tab:full_Real3d_oauroc_results,tab:full_Real3d_pauroc_results}.

\subsection{Results of R3D-AD Instantiation on AnomalyShapeNet}
\label{sec:results_r3dad}

We used the public GitHub implementation of R3D-AD by the authors, but found that in the released training script/configuration, Patch-Gen was not enabled by default during the preparation of training data; as a result, the released training pipeline uses only augmented normal samples (e.g., random rotations) rather than pseudo-anomalies.
To enable a fair comparison, we evaluate three settings for this instantiation: (i) base R3D-AD (no Patch-Gen; exactly as provided), (ii) R3D-AD + Patch-Gen, and (iii) R3D-AD + AF3AD (ours).
We run all three settings on 15 classes of AnomalyShapeNet (index-0 split) and report the results in \Cref{tab:r3dad_results}.
Overall, AF3AD improves performance by 4.5 O-AUROC points over the stronger Patch-Gen baseline and by 8.5 points over base R3D-AD.
Nevertheless, the instantiation based on offset prediction remains stronger and is selected as our main result.
This instantiation mainly shows AF3AD’s modularity across paradigms: object-level detection improves from 62.7 to 67.2 O-AUROC, while P-AUROC stays the same at 60.7. This is likely due to R3D-AD’s downsampling and diffusion-decoder residuals limit point-level localization, regardless of the synthesis method.

\begin{table}[t]
\centering
\caption{Mean performance across 15 AnomalyShapeNet categories (index-0 split) (\%).}
\label{tab:r3dad_results}
\begin{tabular}{lcccc}
\toprule
Method & O-AUROC $\uparrow$ & P-AUROC $\uparrow$ & O-AUPR $\uparrow$ & P-AUPR $\uparrow$ \\
\midrule
R3D-AD  & 58.7 & 59.9 & 63.2 & 7.5 \\
R3D-AD + Patch-Gen & 62.7 & 60.7 & 67.7 & 8.4 \\
R3D-AD + AF3AD & 67.2 & 60.7 & 71.3 & 9.2 \\
\bottomrule
\end{tabular}
\end{table}

\subsection{Ablation Study}

\subsubsection{Effect of Offset Prediction Architecture}
We study the interaction between the offset-head design and the synthesis module under a fixed training budget to verify that AF3AD's gains come from synthesis diversity rather than the modified prediction head.
We compare a baseline head against a multi-head variant for both Norm-AS and AF3AD, and sweep the hidden dimension $h\in\{32,64,128\}$.
The baseline predicts the full 3D offset $(O_x,O_y,O_z)$ using a single MLP, whereas the multi-head design uses separate prediction heads for each coordinate.
\Cref{tab:ablation_arch_synth_asn} shows two consistent trends.
First, the multi-head design improves performance relative to the baseline, with the largest gains appearing under AF3AD training.
Second, AF3AD can underperform Norm-AS at low head capacity (e.g., baseline at $h\in\{32,64\}$), which is expected because AF3AD draws from a larger preset repertoire and therefore reduces per-preset exposure under a fixed budget.
However, once the head is sufficiently expressive, AF3AD becomes competitive in O-AUROC and improves localization, ultimately achieving the best O-AUROC, P-AUROC, and AUPRs at multi-head $h{=}128$.
Crucially, under the matched multi-head $h{=}128$ setting, Norm-AS reaches 91.6/88.9 O/P-AUROC while AF3AD reaches 97.0/93.6, showing that the improvement is not explained by the architectural change alone but requires the diverse synthesis.

\begin{table}[t]
\centering
\scriptsize
\setlength{\tabcolsep}{3.5pt}
\renewcommand{\arraystretch}{1.10}
\caption{Architecture $\times$ synthesis ablation on 15 categories of AnomalyShapeNet (index 0, mean over categories, in \%). Best is in bold and second-best is underlined (per column), computed over the 15-category runs.}
\label{tab:ablation_arch_synth_asn}
\begin{tabular}{llrrrrrr}
\toprule
Synthesis & Offset head & $h$  &
O-AUROC$\uparrow$ & P-AUROC$\uparrow$ & O-AUPR$\uparrow$ & P-AUPR$\uparrow$ \\
\midrule
Norm-AS & Baseline   &  32 & 90.9 & 86.8 & 89.4 & 46.6 \\
Norm-AS & Baseline   &  64 & 91.2 & 87.5 & 91.7 & 49.8 \\
Norm-AS & Baseline   & 128 & 90.5 & 88.8 & 91.4 & 49.3 \\
Norm-AS & Multi-head &  32 & 91.7 & 87.0 & 89.8 & 46.1 \\
Norm-AS & Multi-head &  64 & 91.1 & \underline{90.3} & 92.3 & 50.3 \\
Norm-AS & Multi-head & 128 & 91.6 & 88.9 & 90.7 & 47.1 \\
\midrule
AF3AD   & Baseline   &  32  & 88.4 & 86.2 & 91.0 & 48.1 \\
AF3AD   & Baseline   &  64  & 87.5 & 87.5 & 91.1 & 51.3 \\
AF3AD   & Baseline   & 128  & 90.3 & 87.8 & 91.8 & 50.7 \\
AF3AD   & Multi-head &  32  & 87.1 & 86.4 & 90.0 & 52.4 \\
AF3AD   & Multi-head &  64  & \underline{92.0} & 90.2 & \underline{94.0} & \underline{61.0} \\
AF3AD   & Multi-head & 128  & \textbf{97.0} & \textbf{93.6} & \textbf{97.1} & \textbf{67.7} \\
\bottomrule
\end{tabular}
\end{table}

\subsubsection{Ablation on Pseudo-Anomaly Presets}
To isolate the effect of pseudo-anomaly diversity, we evaluate all combinations of preset groups $\{A, B, C, D\}$ on 15 AnomalyShapeNet categories (index-0 split). 
For each configuration, we enforce equal per-preset exposure by balancing the sampling frequency of the selected presets during training, ensuring a fair comparison across different group combinations.

Overall, performance improves consistently as more groups are combined, with the full set (ABCD) yielding the overall strongest results, while Group B remains the most effective single-group baseline (see \Cref{tab:ablation_onehot_means_checkcross_presetcount_best}).
These findings indicate that maintaining a diverse repertoire of pseudo-anomalies is critical for robust detection and localization.

\begin{table}[t]
\centering
\caption{Preset-group ablation with equal per-preset exposure on AnomalyShapeNet (index-0 split). Mean across categories shared across all configurations (\%).}
\label{tab:ablation_onehot_means_checkcross_presetcount_best}

\renewcommand{\arraystretch}{1.1}
\newcommand{\best}[1]{\textbf{#1}}
\newcommand{\secondd}[1]{\underline{#1}}

\resizebox{\linewidth}{!}{%
\begin{tabular}{@{} lcccc @{\hskip 8pt} | @{\hskip 8pt} lcccc @{}}
\toprule
\textbf{Grp} & O-AUROC & P-AUROC & O-AUPR & P-AUPR & 
\textbf{Grp} & O-AUROC & P-AUROC & O-AUPR & P-AUPR \\
\midrule
A  & 62.5 & 62.4 & 68.0 & 15.3 & BC & 80.0 & 77.6 & 84.0 & 34.0 \\
B  & 74.0 & 72.0 & 79.1 & 27.1 & BD & 83.3 & 80.0 & 87.7 & 37.9 \\
C  & 64.6 & 63.2 & 71.8 & 17.7 & CD & 79.1 & 74.7 & 83.8 & 32.7 \\
D  & 68.3 & 67.9 & 72.8 & 22.2 & ABC& 86.6 & 80.8 & 89.9 & 41.2 \\
AB & 81.5 & 78.5 & 83.9 & 37.7 & ABD& 88.3 & \secondd{84.8} & \secondd{91.6} & \secondd{47.5} \\
AC & 73.8 & 71.5 & 79.3 & 28.1 & ACD& 83.1 & 78.9 & 87.0 & 36.9 \\
AD & 75.7 & 73.4 & 80.4 & 28.8 & BCD& \best{89.0} & 83.7 & 91.4 & 47.3 \\
\multicolumn{5}{c}{} & All& \secondd{88.9} & \best{86.1} & \best{91.6} & \best{53.6} \\
\bottomrule
\end{tabular}%
}
\end{table}

\subsubsection{Robustness to Noisy Data}
We evaluate robustness by adding Gaussian noise ($\sigma \in \{0.001, 0.002, 0.003\}$) to 15 AnomalyShapeNet categories (index-0 split), following~\cite{PO3AD} but with more classes.
Each setting is repeated five times with different random seeds, and we report the mean over runs.
As shown in \Cref{tab:noise_level_means_std}, performance remains largely unchanged under mild perturbations ($\sigma=0.001$), demonstrating robustness to jitter.
While stronger noise gradually reduces accuracy at $\sigma=0.002$ and $\sigma=0.003$, the method retains meaningful detection performance, with the largest impact observed on fine-grained localization (P-AUPR).
\begin{table}[t!]
\centering
\caption{Mean performance (\%, across 15 classes) as a function of Gaussian noise level.}
\label{tab:noise_level_means_std}
\begingroup
\scriptsize
\setlength{\tabcolsep}{4pt}
\renewcommand{\arraystretch}{1.05}
\begin{tabular}{lcccc}
\toprule
$\sigma$ & O-AUROC & P-AUROC & O-AUPR & P-AUPR \\
\midrule
0     & $93.3$ & $97.9$ & $84.8$ & $59.0$ \\
0.001 & $93.3$ & $97.4$ & $84.8$ & $58.0$ \\
0.002 & $82.3$ & $88.0$ & $77.7$ & $34.4$ \\
0.003 & $76.7$ & $80.0$ & $79.7$ & $29.6$ \\
\bottomrule
\end{tabular}
\endgroup
\end{table}

\subsubsection{Empirical Feature-space Alignment}
We further evaluate how well the synthesized anomalies align with real anomalies in feature space using $k$NN-based coverage and Maximum Mean Discrepancy (MMD) on Real3D-AD.
As shown in Table~\ref{tab:real3dad_empirical_coverage}, our synthesis achieves substantially higher coverage of test anomalies while yielding lower MMD than Norm-AS, indicating closer distributional alignment to real anomalous samples.
Although this offline analysis requires access to test anomalies and is therefore not applicable in real deployment settings, it provides additional insight into the quality of the synthesized anomalies. 
Note that the coverage and MMD values are only proxies for the distribution of real anomalies, not proof of physical realism. They suggest that increasing geometric diversity makes synthesized anomalies closer to real anomaly regions, but fully modeling real defects remains an open problem (Further details in \Cref{supp:feature_alignment}).

\begin{table}[t]
\centering
\caption{Real3D-AD anomaly-feature alignment (mean$\pm$std over 12 categories). Original: computed in feature space. UMAP/t-SNE: computed in 2D projections to support \Cref{fig:supp_features}.}
\scriptsize
\setlength{\tabcolsep}{3.5pt}
\renewcommand{\arraystretch}{1.12}
\resizebox{\columnwidth}{!}{%
\begin{tabular}{lcc cc cc}
\toprule
& \multicolumn{2}{c}{Original (feature space)} & \multicolumn{2}{c}{UMAP (2D)} & \multicolumn{2}{c}{t-SNE (2D)} \\
\cmidrule(lr){2-3}\cmidrule(lr){4-5}\cmidrule(lr){6-7}
Method
& $\mathrm{Cov}(\mathrm{T}\!\rightarrow\!\cdot)\uparrow$
& $\mathrm{MMD}\downarrow$
& Hull IoU$\uparrow$
& $\mathrm{Cov}(\mathrm{T}\!\rightarrow\!\cdot)\uparrow$
& Hull IoU$\uparrow$
& $\mathrm{Cov}(\mathrm{T}\!\rightarrow\!\cdot)\uparrow$ \\
\midrule
Norm-AS
& 0.087$\pm$0.130 & 0.326$\pm$0.146
& 0.157$\pm$0.133 & 0.007$\pm$0.013
& 0.191$\pm$0.105 & 0.002$\pm$0.006 \\
Ours
& \textbf{0.380$\pm$0.187} & \textbf{0.105$\pm$0.079}
& \textbf{0.483$\pm$0.242} & \textbf{0.242$\pm$0.107}
& \textbf{0.521$\pm$0.220} & \textbf{0.196$\pm$0.118} \\
\bottomrule
\end{tabular}}
\label{tab:real3dad_empirical_coverage}
\end{table}

\subsection{Qualitative Results}
\begin{figure}[t]
    \centering
    \includegraphics[width=0.9\textwidth]{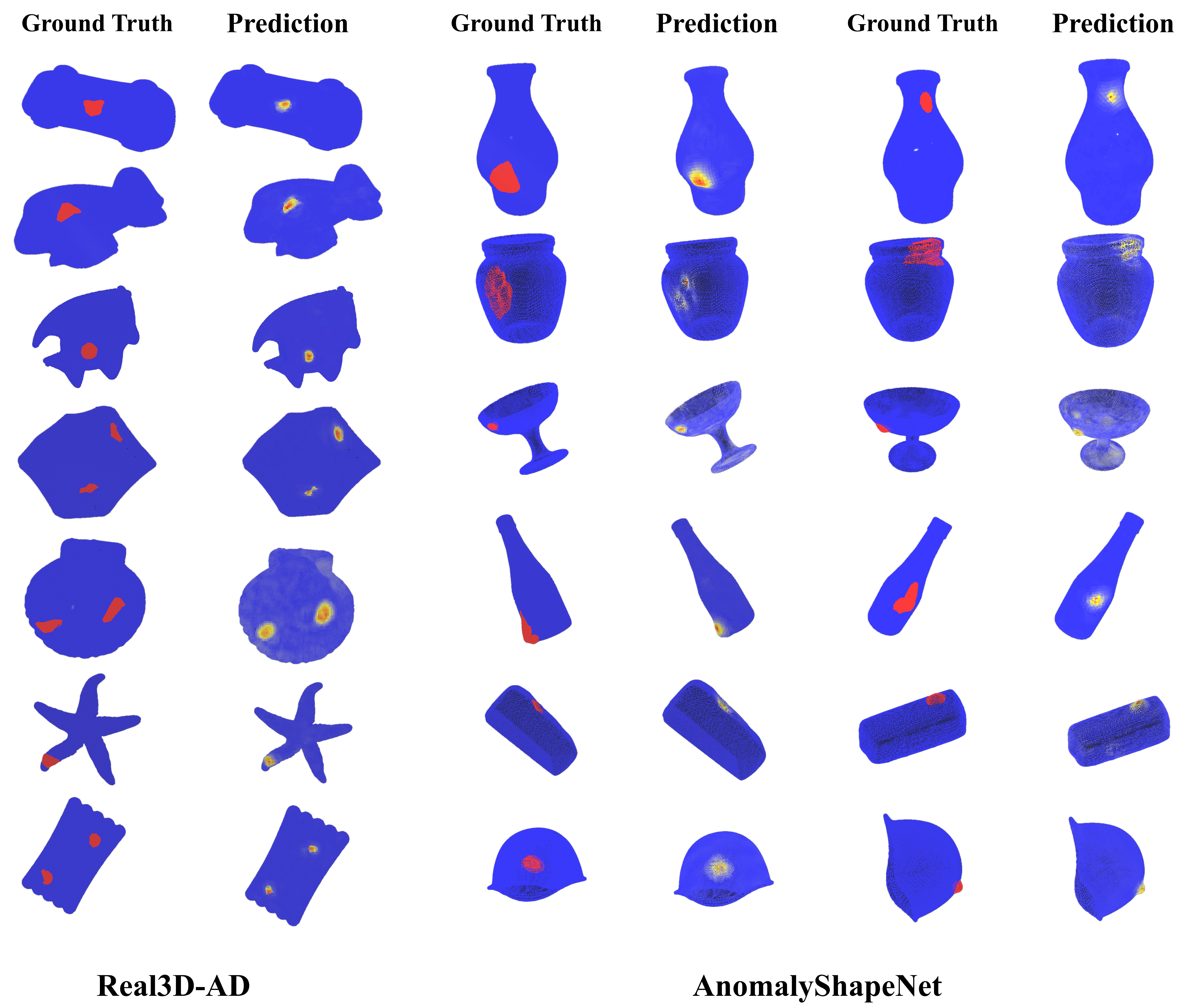}
    \caption{Samples of anomaly prediction using offset prediction method trained with AF3AD.}
    \label{fig:sample_outputs}
\end{figure}

\Cref{fig:sample_outputs} displays anomaly score maps that illustrate the effectiveness of our localization method on various test samples from the Real3D-AD and AnomalyShapeNet datasets.
Our approach effectively identifies and highlights anomalous regions while assigning appropriate scores to normal points, a critical requirement in the 3DAD task.
However, we observe a few failure cases, which are detailed in Supplementary \Cref{fig:bad_results}.

\section{Discussion}
AF3AD functions as a modular synthesis framework capable of generating diverse pseudo-anomalies to train detection methods that rely on synthetic data.
While we demonstrate integration with two different anomaly detection methods, the framework's modular design enables adoption by other paradigms.
Our kernel-based synthesis can easily replace simplistic generators, such as PO3AD's Norm-AS or R3D-AD's Patch-Gen.
Across both benchmarks, AF3AD improves over competing methods at the object level and stays competitive at the point level, with all gains averaged over three seeds (\Cref{sec:results}).
We observe that our "Group A" presets are similar to the anomalies generated by Norm-AS, the synthesis module used in PO3AD.
As shown in \Cref{tab:ablation_onehot_means_checkcross_presetcount_best}, integrating additional types of anomalies beyond this baseline consistently leads to better results.
These gains persist under realistic perturbations, such as random noise (\Cref{tab:noise_level_means_std}), demonstrating the robustness of our proposed framework.
In \Cref{tab:real3dad_empirical_coverage}, we empirically show that utilizing diverse anomaly presets allows us to populate wider regions of the "anomaly space" by considering factors like scale, anisotropy, boundary sharpness, displacement ratio, and affected area.
This prevents the model from overfitting to a narrow range of synthetic anomalies.
Furthermore, our approach enables real-time, controllable anomaly generation without requiring labeled data.
This diversity acts as an inductive bias, helping to train a more powerful anomaly detector—a benefit also observed in 2D anomaly detection studies \cite{10.1145/3627673.3679623}. 
AF3AD has two main limitations. First, because it synthesizes anomalies
through smooth geometric displacement of points, its coverage is uneven
across real defect types: it directly covers geometric surface
deformations such as bulges, dents, ridges, trenches, and shear
(\Cref{tab:anomaly_taxonomy}); it only partially covers sharp or
high-frequency defects such as cracks and fine scratches, which smooth
kernels approximate but do not reproduce exactly; and it does not address
non-geometric defects such as material flaws or purely appearance-based
anomalies, since the synthesis modifies geometry rather than surface
appearance. Second, the preset configurations are fixed and not tailored
to a given dataset, so if the synthesized anomalies diverge too far from
the real defect distribution, generalization may suffer. Incorporating
minimal domain guidance and developing a principled way to predict which
anomaly types are most effective for a specific dataset are therefore
promising directions for future work.
\section{Conclusion}
In this work, we proposed AF3AD, a modular framework for synthesizing diverse pseudo-anomalies to generate training signals for 3D anomaly detection. Our framework provides a systematic taxonomy of geometric deformations with explicit parametric control, operating independently of specific detection architectures. We demonstrated its effectiveness by integrating AF3AD with an offset-prediction detector, achieving superior performance on AnomalyShapeNet (91.5\% O-AUROC) and Real3D-AD (85.2\% O-AUROC), with gains of 5.5 and 7.1 points over the previous best methods, respectively. Extensive ablations confirm that diverse synthesis strategies consistently outperform limited baselines and remain robust under noise. Future work will explore integrating generative modeling for richer anomaly representations and implementing curriculum learning to progressively introduce complex anomalies.

%
%
\newpage
\bibliographystyle{splncs04}
\bibliography{main}

\begin{thebibliography}{10}
\providecommand{\url}[1]{\texttt{#1}}
\providecommand{\urlprefix}{URL }
\providecommand{\doi}[1]{https://doi.org/#1}

\bibitem{MVTec3D}
Bergmann, P., Jin, X., Sattlegger, D., Steger, C.: The mvtec 3d-ad dataset for unsupervised 3d anomaly detection and localization. CoRR  \textbf{abs/2112.09045} (2021), \url{https://arxiv.org/abs/2112.09045}

\bibitem{CPMF}
Cao, Y., Xu, X., Shen, W.: Complementary pseudo multimodal feature for point cloud anomaly detection. Pattern Recognit.  \textbf{156},  110761 (2024). \doi{10.1016/J.PATCOG.2024.110761}, \url{https://doi.org/10.1016/j.patcog.2024.110761}

\bibitem{cao2024surveyvisualanomalydetection}
Cao, Y., Xu, X., Zhang, J., Cheng, Y., Huang, X., Pang, G., Shen, W.: A survey on visual anomaly detection: Challenge, approach, and prospect. CoRR  \textbf{abs/2401.16402} (2024). \doi{10.48550/ARXIV.2401.16402}, \url{https://doi.org/10.48550/arXiv.2401.16402}

\bibitem{ShapeNet}
Chang, A.X., Funkhouser, T.A., Guibas, L.J., Hanrahan, P., Huang, Q., Li, Z., Savarese, S., Savva, M., Song, S., Su, H., Xiao, J., Yi, L., Yu, F.: Shapenet: An information-rich 3d model repository. CoRR  \textbf{abs/1512.03012} (2015), \url{http://arxiv.org/abs/1512.03012}

\bibitem{glfm}
Cheng, Y., Cao, Y., Wang, D., Shen, W., Li, W.: Boosting global-local feature matching via anomaly synthesis for multi-class point cloud anomaly detection. {IEEE} Trans Autom. Sci. Eng.  \textbf{22},  12560--12571 (2025). \doi{10.1109/TASE.2025.3544462}, \url{https://doi.org/10.1109/TASE.2025.3544462}

\bibitem{MiniShift}
Cheng, Y., Sun, Y., Zhang, H., Shen, W., Cao, Y.: Towards high-resolution 3d anomaly detection: {A} scalable dataset and real-time framework for subtle industrial defects. In: Koenig, S., Jenkins, C., Taylor, M.E. (eds.) Fortieth {AAAI} Conference on Artificial Intelligence, Thirty-Eighth Conference on Innovative Applications of Artificial Intelligence, Sixteenth Symposium on Educational Advances in Artificial Intelligence, {AAAI} 2026, Singapore, January 20-27, 2026. pp. 3327--3334. {AAAI} Press (2026). \doi{10.1609/AAAI.V40I5.37328}, \url{https://doi.org/10.1609/aaai.v40i5.37328}

\bibitem{mink}
Choy, C.B., Gwak, J., Savarese, S.: 4d spatio-temporal convnets: Minkowski convolutional neural networks. In: {IEEE} Conference on Computer Vision and Pattern Recognition, {CVPR} 2019, Long Beach, CA, USA, June 16-20, 2019. pp. 3075--3084. Computer Vision Foundation / {IEEE} (2019). \doi{10.1109/CVPR.2019.00319}, \url{http://openaccess.thecvf.com/content\_CVPR\_2019/html/Choy\_4D\_Spatio-Temporal\_ConvNets\_Minkowski\_Convolutional\_Neural\_Networks\_CVPR\_2019\_paper.html}

\bibitem{adrd}
Deng, H., Li, X.: Anomaly detection via reverse distillation from one-class embedding. In: {IEEE/CVF} Conference on Computer Vision and Pattern Recognition, {CVPR} 2022, New Orleans, LA, USA, June 18-24, 2022. pp. 9727--9736. {IEEE} (2022). \doi{10.1109/CVPR52688.2022.00951}, \url{https://doi.org/10.1109/CVPR52688.2022.00951}

\bibitem{3DAD_manufacturing_survey}
Du, J., Tao, C., Cao, X., Tsung, F.: 3d vision-based anomaly detection in manufacturing: A survey. Frontiers of Engineering Management  \textbf{12}(2),  343--360 (Jun 2025). \doi{10.1007/s42524-025-4189-9}, \url{https://doi.org/10.1007/s42524-025-4189-9}

\bibitem{cflow-ad}
Gudovskiy, D.A., Ishizaka, S., Kozuka, K.: {CFLOW-AD:} real-time unsupervised anomaly detection with localization via conditional normalizing flows. In: {IEEE/CVF} Winter Conference on Applications of Computer Vision, {WACV} 2022, Waikoloa, HI, USA, January 3-8, 2022. pp. 1819--1828. {IEEE} (2022). \doi{10.1109/WACV51458.2022.00188}, \url{https://doi.org/10.1109/WACV51458.2022.00188}

\bibitem{BTF}
Horwitz, E., Hoshen, Y.: Back to the feature: Classical 3d features are (almost) all you need for 3d anomaly detection. In: {IEEE/CVF} Conference on Computer Vision and Pattern Recognition, {CVPR} 2023 - Workshops, Vancouver, BC, Canada, June 17-24, 2023. pp. 2968--2977. {IEEE} (2023). \doi{10.1109/CVPRW59228.2023.00298}, \url{https://doi.org/10.1109/CVPRW59228.2023.00298}

\bibitem{DL_Advancements_in_AD}
Huang, H., Wang, P., Pei, J., Wang, J., Alexanian, S., Niyato, D.: Deep learning advancements in anomaly detection: {A} comprehensive survey. {IEEE} Internet Things J.  \textbf{12}(21),  44318--44342 (2025). \doi{10.1109/JIOT.2025.3585884}, \url{https://doi.org/10.1109/JIOT.2025.3585884}

\bibitem{pca}
Jolliffe, I.T.: Principal Component Analysis. Springer Series in Statistics, Springer (1986). \doi{10.1007/978-1-4757-1904-8}, \url{https://doi.org/10.1007/978-1-4757-1904-8}

\bibitem{10.1145/3627673.3679623}
Kim, H., Lee, C.: Enhancing anomaly detection via generating diversified and hard-to-distinguish synthetic anomalies. In: Proceedings of the 33rd ACM International Conference on Information and Knowledge Management. p. 1089–1098. CIKM '24, Association for Computing Machinery, New York, NY, USA (2024). \doi{10.1145/3627673.3679623}, \url{https://doi.org/10.1145/3627673.3679623}

\bibitem{AnomalyShapeNet}
Li, W., Xu, X., Gu, Y., Zheng, B., Gao, S., Wu, Y.: Towards scalable 3d anomaly detection and localization: {A} benchmark via 3d anomaly synthesis and {A} self-supervised learning network. In: {IEEE/CVF} Conference on Computer Vision and Pattern Recognition, {CVPR} 2024, Seattle, WA, USA, June 16-22, 2024. pp. 22207--22216. {IEEE} (2024). \doi{10.1109/CVPR52733.2024.02096}, \url{https://doi.org/10.1109/CVPR52733.2024.02096}

\bibitem{Liang2024LookIF}
Liang, H., Xie, G., Hou, C., Wang, B., Gao, C., Wang, J.: Look inside for more: Internal spatial modality perception for 3d anomaly detection. In: AAAI Conference on Artificial Intelligence (2024). \doi{10.48550/arXiv.2412.13461}

\bibitem{uiad_survey}
Lin, Y., Chang, Y., Tong, X., Yu, J., Liotta, A., Huang, G., Song, W., Zeng, D., Wu, Z., Wang, Y., Zhang, W.: A survey on rgb, 3d, and multimodal approaches for unsupervised industrial image anomaly detection. Inf. Fusion  \textbf{121},  103139 (2025). \doi{10.1016/J.INFFUS.2025.103139}, \url{https://doi.org/10.1016/j.inffus.2025.103139}

\bibitem{Real3DAD}
Liu, J., Xie, G., Chen, R., Li, X., Wang, J., Liu, Y., Wang, C., Zheng, F.: Real3d-ad: {A} dataset of point cloud anomaly detection. In: Oh, A., Naumann, T., Globerson, A., Saenko, K., Hardt, M., Levine, S. (eds.) Advances in Neural Information Processing Systems 36: Annual Conference on Neural Information Processing Systems 2023, NeurIPS 2023, New Orleans, LA, USA, December 10 - 16, 2023 (2023), \url{http://papers.nips.cc/paper\_files/paper/2023/hash/611b896d447df43c898062358df4c114-Abstract-Datasets\_and\_Benchmarks.html}

\bibitem{PointMAE}
Pang, Y., Wang, W., Tay, F.E.H., Liu, W., Tian, Y., Yuan, L.: Masked autoencoders for point cloud self-supervised learning. In: Avidan, S., Brostow, G.J., Ciss{\'{e}}, M., Farinella, G.M., Hassner, T. (eds.) Computer Vision - {ECCV} 2022 - 17th European Conference, Tel Aviv, Israel, October 23-27, 2022, Proceedings, Part {II}. Lecture Notes in Computer Science, vol. 13662, pp. 604--621. Springer (2022). \doi{10.1007/978-3-031-20086-1\_35}, \url{https://doi.org/10.1007/978-3-031-20086-1\_35}

\bibitem{tsad}
Qin, J., Gu, C., Yu, J., Zhang, C.: Teacher-student network for 3d point cloud anomaly detection with few normal samples. Expert Syst. Appl.  \textbf{228},  120371 (2023). \doi{10.1016/J.ESWA.2023.120371}, \url{https://doi.org/10.1016/j.eswa.2023.120371}

\bibitem{patchcore}
Roth, K., Pemula, L., Zepeda, J., Sch{\"{o}}lkopf, B., Brox, T., Gehler, P.V.: Towards total recall in industrial anomaly detection. In: {IEEE/CVF} Conference on Computer Vision and Pattern Recognition, {CVPR} 2022, New Orleans, LA, USA, June 18-24, 2022. pp. 14298--14308. {IEEE} (2022). \doi{10.1109/CVPR52688.2022.01392}, \url{https://doi.org/10.1109/CVPR52688.2022.01392}

\bibitem{M3DM}
Wang, Y., Peng, J., Zhang, J., Yi, R., Wang, Y., Wang, C.: Multimodal industrial anomaly detection via hybrid fusion. In: {IEEE/CVF} Conference on Computer Vision and Pattern Recognition, {CVPR} 2023, Vancouver, BC, Canada, June 17-24, 2023. pp. 8032--8041. {IEEE} (2023). \doi{10.1109/CVPR52729.2023.00776}, \url{https://doi.org/10.1109/CVPR52729.2023.00776}

\bibitem{DFR}
Yang, J., Shi, Y., Qi, Z.: {DFR:} deep feature reconstruction for unsupervised anomaly segmentation. CoRR  \textbf{abs/2012.07122} (2020), \url{https://arxiv.org/abs/2012.07122}

\bibitem{PO3AD}
Ye, J., Zhao, W., Yang, X., Cheng, G., Huang, K.: {PO3AD:} predicting point offsets toward better 3d point cloud anomaly detection. In: {IEEE/CVF} Conference on Computer Vision and Pattern Recognition, {CVPR} 2025, Nashville, TN, USA, June 11-15, 2025. pp. 1353--1362. Computer Vision Foundation / {IEEE} (2025). \doi{10.1109/CVPR52734.2025.00134}, \url{https://openaccess.thecvf.com/content/CVPR2025/html/Ye\_PO3AD\_Predicting\_Point\_Offsets\_toward\_Better\_3D\_Point\_Cloud\_Anomaly\_CVPR\_2025\_paper.html}

\bibitem{pointbert}
Yu, X., Tang, L., Rao, Y., Huang, T., Zhou, J., Lu, J.: Point-bert: Pre-training 3d point cloud transformers with masked point modeling. In: {IEEE/CVF} Conference on Computer Vision and Pattern Recognition, {CVPR} 2022, New Orleans, LA, USA, June 18-24, 2022. pp. 19291--19300. {IEEE} (2022). \doi{10.1109/CVPR52688.2022.01871}, \url{https://doi.org/10.1109/CVPR52688.2022.01871}

\bibitem{Reg2Inv}
Yu, Y., Chen, Z., Xu, X., Zhang, L., Yang, H., Nie, Y., He, S.: Registration is a powerful rotation-invariance learner for 3d anomaly detection. CoRR  \textbf{abs/2510.16865} (2025). \doi{10.48550/ARXIV.2510.16865}, \url{https://doi.org/10.48550/arXiv.2510.16865}

\bibitem{R3DAD}
Zhou, Z., Wang, L., Fang, N., Wang, Z., Qiu, L., Zhang, S.: {R3D-AD:} reconstruction via diffusion for 3d anomaly detection. In: Leonardis, A., Ricci, E., Roth, S., Russakovsky, O., Sattler, T., Varol, G. (eds.) Computer Vision - {ECCV} 2024 - 18th European Conference, Milan, Italy, September 29-October 4, 2024, Proceedings, Part {XXXVI}. Lecture Notes in Computer Science, vol. 15094, pp. 91--107. Springer (2024). \doi{10.1007/978-3-031-72764-1\_6}, \url{https://doi.org/10.1007/978-3-031-72764-1\_6}

\bibitem{Group3AD}
Zhu, H., Xie, G., Hou, C., Dai, T., Gao, C., Wang, J., Shen, L.: Towards high-resolution 3d anomaly detection via group-level feature contrastive learning. In: Cai, J., Kankanhalli, M.S., Prabhakaran, B., Boll, S., Subramanian, R., Zheng, L., Singh, V.K., C{\'{e}}sar, P., Xie, L., Xu, D. (eds.) Proceedings of the 32nd {ACM} International Conference on Multimedia, {MM} 2024, Melbourne, VIC, Australia, 28 October 2024 - 1 November 2024. pp. 4680--4689. {ACM} (2024). \doi{10.1145/3664647.3680919}, \url{https://doi.org/10.1145/3664647.3680919}

\bibitem{clip3d-ad}
Zuo, Z., Dong, J., Wu, Y., Qu, Y., Wu, Z.: {CLIP3D-AD:} extending {CLIP} for 3d few-shot anomaly detection with multi-view images generation. CoRR  \textbf{abs/2406.18941} (2024). \doi{10.48550/ARXIV.2406.18941}, \url{https://doi.org/10.48550/arXiv.2406.18941}

\end{thebibliography}

\newpage

\appendix

\renewcommand{\thefigure}{S\arabic{figure}}
\renewcommand{\thesection}{S\arabic{section}}
\renewcommand{\thesubsection}{S\arabic{subsection}}
\renewcommand{\thetable}{S\arabic{table}}
\setcounter{figure}{0}
\setcounter{table}{0}

\section*{Supplementary Materials}
\subsection{Offset-Prediction Detector: Architectural Details}
\label{supp:offset_detector}

This section provides detailed architectural details and training procedures for our offset-prediction detector instantiation, complementing the overview in Section~\ref{subsection:offset_pred} of the main paper.
We modified the architecture of PO3AD to better suit our diverse pseudo-anomaly synthesis by making it more capable of learning more diverse structures. While our version uses the same point feature encoder $F$ (MinkUNet34C), the architecture of the prediction head is 
altered in two key ways:

\subsubsection{Multi-head prediction.} Instead of predicting all three coordinates through a single MLP, our version employs separate prediction heads for the $x$, $y$, and $z$ offsets. 
Each head features a larger hidden dimension (128 vs. 32 in baseline), allowing it to specialize in a specific coordinate direction. This design choice enables better learning of directional deformations, particularly beneficial for 
our diverse preset taxonomy that includes both normal and tangential anomalies.

Additionally, $L_1$ regularization is incorporated to improve the stability of the training process when learning from diverse anomaly types.
This prevents overfitting to specific deformation patterns and encourages sparse, interpretable offset predictions.

The relationship between the input pseudo-anomalous point cloud $P'$, the feature encoder $F$, and the offset predictor $OP$ is defined as follows:
\begin{equation}
    OP(F(P')) \rightarrow (O_x, O_y, O_z)
\end{equation}

\subsubsection{Training Objective}

To train the offset predictor, we utilize the following loss function, which follows the methodology of PO3AD with the addition of $L_1$ regularization:
\begin{align}
\mathcal{L}_{off} &= \mathcal{L}_{dist} + \mathcal{L}_{dir} + \lambda \sum_{j} |w_j|, \\
\mathcal{L}_{dist} &= \frac{1}{N} \sum_{i=1}^{N} \left\| o_i^{prd} - o_i^{gt} \right\|_{o_i^{prd} \in O^{prd}, o_i^{gt} \in O^{gt}}, \\
\mathcal{L}_{dir} &= - \frac{1}{N} \sum_{i=1}^{N} \frac{o_i^{prd}}{\| o_i^{prd} \|_2 + \epsilon} \cdot \frac{o_i^{gt}}{\| o_i^{gt} \|_2 + \epsilon} \bigg|_{o_i^{gt} \in O^{gt}, o_i^{prd} \in O^{prd}}
\end{align}

where $\mathcal{L}_{dist}$ measures the Euclidean distance between predicted and ground-truth offsets, $\mathcal{L}_{dir}$ ensures directional alignment through cosine similarity, and the $L_1$ term (weighted by $\lambda$) regularizes the network weights $w_j$. The ground-truth offsets $o_i^{gt}$ are computed as the difference between deformed and original point positions: $o_i^{gt} = p_i' - p_i$, where $p_i'$ and $p_i$ are from the pseudo-anomalous and normal point clouds, respectively.

\subsection{Down-sampling Effect on Real3D-AD}
\label{subsubsection:ablation_downsampling}
\Cref{tab:real3dad_downsample_ablation_summary} evaluates various down-sampling strategies on Real3D-AD to balance accuracy and efficiency.
We compare three settings: \texttt{None}, which uses the full point cloud without down-sampling; \texttt{random(X)}, which randomly retains a fraction \(X\) of the input points; and \texttt{voxel+fps}, which first applies voxel grid sampling and then uses Farthest Point Sampling to obtain a fixed number of points.
While the full point cloud yields the highest P-AUROC, it requires $7.1$\, hours of training.
We selected the \texttt{voxel+fps(15k)} configuration, as it achieves the best O-AUROC ($70.9\%$) and O-AUPR ($72.2\%$) while reducing training time to $5.2$\,hours.
This choice provides an optimal trade-off between detection performance and computational cost.

\begin{table}[t]
\centering
\caption{Ablation results on different down-sampling methods on 5 classes of Real3D-AD. Metrics are reported as percentages; time is in hours. Best is bold; second-best is underlined.}
\label{tab:real3dad_downsample_ablation_summary}
\begingroup
\setlength{\tabcolsep}{3pt}
\renewcommand{\arraystretch}{1.08}
\scriptsize
\resizebox{\linewidth}{!}{%
\begin{tabular}{lccccc}
\toprule
\textbf{Config} & \textbf{O-AUROC}~$\uparrow$ & \textbf{P-AUROC}~$\uparrow$ & \textbf{O-AUPR}~$\uparrow$ & \textbf{P-AUPR}~$\uparrow$ & \textbf{Time (h)}~$\downarrow$ \\
\midrule
none(full)        & 66.3 & \textbf{78.7} & 69.0 & \underline{23.1} & 7.1 \\
random(0.3)       & \underline{70.2} & 74.0 & \underline{70.9} & 20.2 & \textbf{4.3} \\
random(0.5)       & 67.1 & 73.4 & 69.5 & 20.4 & 5.2 \\
random(0.7)       & 67.2 & \underline{78.4} & 68.8 & \textbf{24.3} & 6.0 \\
voxel+fps(10k)    & 67.8 & 75.0 & 69.7 & 19.3 & \underline{4.6} \\
voxel+fps(15k)    & \textbf{70.9} & 76.2 & \textbf{72.2} & 21.4 & 5.2 \\
voxel+fps(20k)    & 69.2 & 77.0 & 69.9 & 21.0 & 5.8 \\
\bottomrule
\end{tabular}%
}
\endgroup
\end{table}

\subsection{Feature-Space Alignment Analysis}
\label{supp:feature_alignment}

\begin{figure}[h!]
    \centering
    \includegraphics[width=\textwidth]{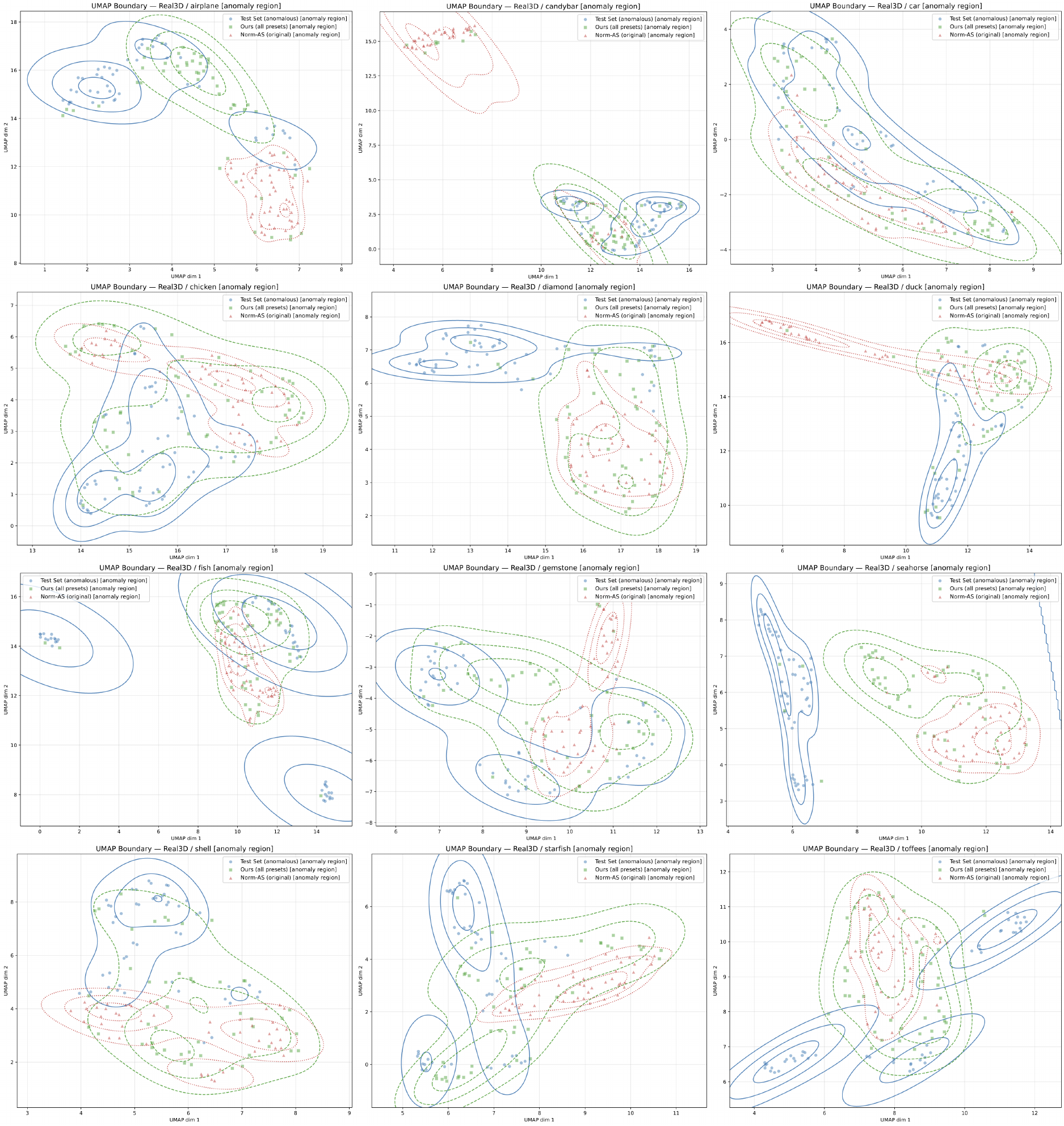}
    \caption{UMAP visualization on all categories of Real3D-AD. Blue: test anomalies, Green: our 
    synthesis, Red: Norm-AS.}
    \label{fig:supp_features}
\end{figure}

We extract features from anomalous regions using the MinkUNet34C point cloud feature encoder for Real3D-AD test anomalies, our synthesized anomalies, and Norm-AS anomalies. We measure bidirectional $k$NN coverage ($k=5$) and distribution distances (MMD, Wasserstein).
Figure~\ref{fig:supp_features} shows UMAP visualization for all categories of Real3D-AD. 
Our synthesis achieves higher coverage of test anomalies compared to Norm-AS, indicating better distributional alignment. 

\begin{figure}[h!]
    \centering
    \includegraphics[width=0.7\textwidth]{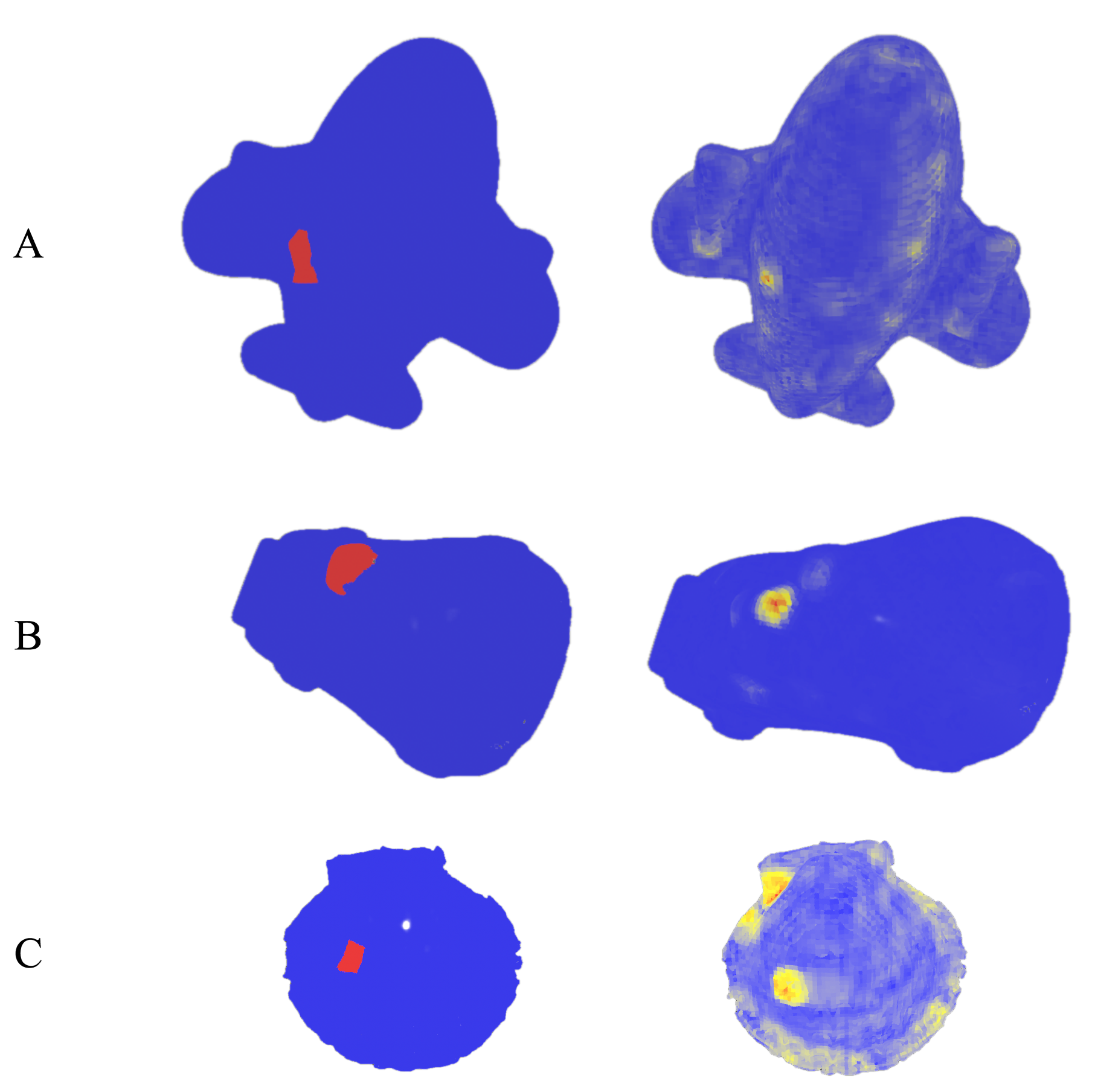}
    \caption{Failures of the anomaly detection module in detail. (A) Some normal parts of the point cloud have high anomaly scores (shown in yellow). (B) The anomaly location is predicted incorrectly. (C) Similar to (A), normal regions receive higher anomaly scores than the true anomalous regions.}
    \label{fig:bad_results}
\end{figure}

\begin{table}[t]
    \centering
    \caption{Pseudo-anomaly preset configurations. Unless noted, each preset uses the global radius $R$.}
    \label{tab:anomaly_presets}
    \begin{threeparttable}
    \small
    \setlength{\tabcolsep}{3pt}
    \renewcommand{\arraystretch}{1.16}
    \begin{tabularx}{\linewidth}{@{} l c c c X @{}}
        \toprule
        \textbf{Preset} & \makecell{\textbf{Radii}\\$(r_u,r_v,r_n)$} & \textbf{K} & \textbf{Dir.} & \textbf{Controls and parameters} \\
        \midrule
        Basic Bulge/Dent & $(1.0,1.0,1.0)$ & C & $\bar{\mathbf{n}}$ & one-sided; $s=\pm1$; $B=1.00$; $\sigma=0.35$ \\
        Ridge & $(2.5,0.7,0.6)$ & G & $\bar{\mathbf{n}}$ & global-axis gated; one-sided; $s=+1$; $B=1.00$; $\sigma=0.4$, $\mathbf{n}_g=(0,0,1)$ \\
        Trench & $(3.0,0.7,0.5)$ & C & $\bar{\mathbf{n}}$ & global-axis gated; one-sided; $s=-1$; $B=1.00$; $\mathbf{n}_g=(0,0,1)$ \\
        Flat Spot & $(1.3,1.0,0.5)$ & G & $\mathbf{n}_i$ & normal-gated; one-sided; $s=-1$; $B=1.00$; $\sigma=0.5$ \\
        Skewed Crater & $(1.0,1.0,0.6)$ & C & $\mathbf{n}_i$ & global-axis gated; one-sided; $s=-1$; $B=1.00$; $\mathbf{n}_g\approx(0.2,0.3,0.9)$ \\
        Shear ($u/v$) & $(1.2,1.2,0.8)$ & C & $\mathbf{t}_u/\mathbf{t}_v$ & no gating; two-sided; $s=\pm1$; $B=1.00$ \\
        Double-sided Ripple & $(1.0,1.0,0.8)$ & C & $\bar{\mathbf{n}}$ & no gating; two-sided; random sign; $B=1.00$; Bernoulli \\
        Micro Dimple & $(0.4,0.4,0.4)$ & C & $\bar{\mathbf{n}}$ & normal-gated; one-sided; $s=-1$; $B=0.25$; $\sigma=0.4$; $\times N$ random locations; no global $R$ \\
        Drag/Stretch & $(2.0,0.8,0.6)$ & G & $\mathbf{t}_u$ & global-axis gated; one-sided; $s=+1$; $B=0.17$; $\sigma=0.5$, $\mathbf{n}_g=(0,1,0)$ \\
        \bottomrule
    \end{tabularx}
    \begin{tablenotes}[flushleft]
        \footnotesize
        \item K: kernel (C: cosine; G: Gaussian). $\bar{\mathbf{n}}$: mean normal; $\mathbf{n}_i$: point normal; $\mathbf{t}_u/\mathbf{t}_v$: tangent directions. $s$ denotes displacement sign, and $B$ is the scale factor relative to the base magnitude.
    \end{tablenotes}
    \end{threeparttable}
\end{table}

\subsection{Training Data Preprocessing on Real3D-AD (Domain Shift Ablation)}
\label{app:real3dad_train_data_ablation}

Real3D-AD exhibits a noticeable train--test domain shift due to differences in scanning completeness and boundary artifacts (e.g., partial views and cut edges). 
Because the literature does not provide a clear, standardized preprocessing protocol for training on Real3D-AD, we conduct an ablation study over multiple training data compositions and cut-style augmentations to justify our final choice. 
For computational tractability, we run this study on a subset of \textbf{five categories} (airplane, car, chicken, diamond, gemstone).
We evaluate ten settings that vary whether training uses full scans (\texttt{full}), cut scans (\texttt{cut}), their combination (\texttt{cut} + \texttt{full}), and edge-dropout augmentation.
The results in \Cref{tab:real3dad_train_data_ablation} show that incorporating cut-style data generally improves performance compared to training on uncut full scans.
Although \texttt{cut} + \texttt{full} variants achieve the best overall scores, training on \texttt{cut} only performs comparably to the top-performing configurations, supporting our decision to use the \texttt{cut} preprocessing in the main experiments on Real3D-AD as a simple and effective choice.

\begin{table}[t]
\centering
\caption{Ablation study on Real3D-AD training data preprocessing to mitigate train--test domain shift. Experiments are conducted on five categories: airplane, car, chicken, diamond, and gemstone. We report the mean object-level (O-AUROC) and point-level (P-AUROC), along with their average. The best and second-best results are highlighted (Bold: Best, Underline: Second-best).}
\label{tab:real3dad_train_data_ablation}
\resizebox{\linewidth}{!}{
\begin{tabular}{lccc}
\toprule
Training Data Preprocessing & O-AUROC $\uparrow$ & P-AUROC $\uparrow$ & (O+P)/2 $\uparrow$ \\
\midrule
\texttt{full} (no plane cut)                          & 60.9 & 66.8 & 63.9 \\
\texttt{full} + plane cut (default)                   & 65.6 & 70.3 & 68.0 \\
\texttt{full} + plane cut (aggressive)                & 59.3 & 67.2 & 63.2 \\
\texttt{full} + plane cut (gentle)                    & 64.9 & 73.1 & 69.0 \\
\texttt{cut}                                          & \underline{72.0} & \underline{81.7} & \underline{76.9} \\
\texttt{cut} + \texttt{full} (no plane cut)           & \textbf{73.7} & \textbf{82.0} & \textbf{77.9} \\
\texttt{cut} + \texttt{full} + plane cut (default)    & 72.9 & 80.6 & 76.8 \\
\texttt{cut} + edge dropout                           & 73.1 & 79.5 & 76.3 \\
\texttt{cut} + \texttt{full} + edge dropout (no plane cut) & 73.0 & 79.9 & 76.5 \\
\texttt{cut} + \texttt{full} + plane cut (default) + edge dropout & 69.2 & 77.8 & 73.5 \\
\bottomrule
\end{tabular}
}
\end{table}

\newcommand{\topone}[1]{\textbf{#1}}      
\newcommand{\toptwo}[1]{\underline{#1}}   

\begin{table*}[tbp]
\centering
\caption{Comparison of O-AUROC results (\%) at the object level. Bold: Best, Underline: Second-best.}
\label{tab:full_ANS_oauroc_results}
\footnotesize
\setlength{\tabcolsep}{1.5pt}
\renewcommand{\arraystretch}{1.12}
\resizebox{\textwidth}{!}{%
\begin{tabular}{l cccccccccccccccccccc}
\toprule
\multicolumn{21}{c}{O-AUROC($\uparrow$)} \\ \midrule
Method & Ashtray0 & Bag0 & Bottle0 & Bottle1 & Bottle3 & Bowl0 & Bowl1 & Bowl2 & Bowl3 & Bowl4 & Bowl5 & Bucket0 & Bucket1 & Cap0 & Cap3 & Cap4 & Cap5 & Cup0 & Cup1 & Eraser0 \\ \midrule
BTF(RAW) & 57.8 & 41.0 & 59.7 & 51.0 & 56.8 & 56.4 & 26.4 & 52.5 & 38.5 & 66.4 & 41.7 & 61.7 & 32.1 & 66.8 & 52.7 & 46.8 & 37.3 & 40.3 & 52.1 & 52.5 \\
BTF(FPFH) & 42.0 & 54.6 & 34.4 & 54.6 & 32.2 & 50.9 & 66.8 & 51.0 & 49.0 & 60.9 & 69.9 & 40.1 & 63.3 & 61.8 & 52.2 & 52.0 & 58.6 & 58.6 & 61.0 & 71.9 \\
M3DM & 57.7 & 53.7 & 57.4 & 63.7 & 54.1 & 63.4 & 66.3 & 68.4 & 61.7 & 46.4 & 40.9 & 30.9 & 50.1 & 55.7 & 42.3 & 77.7 & 63.9 & 53.9 & 55.6 & 62.7 \\
PatchCore(FPFH) & 58.7 & 57.1 & 60.4 & 66.7 & 57.2 & 50.4 & 63.9 & 61.5 & 53.7 & 49.4 & 55.8 & 46.9 & 55.1 & 58.0 & 45.3 & 75.7 & 79.0 & 60.0 & 58.6 & 65.7 \\
PatchCore(PMAE) & 59.1 & 60.1 & 51.3 & 60.1 & 65.0 & 52.3 & 62.9 & 45.8 & 57.9 & 50.1 & 59.3 & 59.3 & 56.1 & 58.9 & 47.6 & 72.7 & 53.8 & 61.0 & 55.6 & 67.7 \\
CPMF & 35.3 & 64.3 & 52.0 & 48.2 & 40.5 & 78.3 & 63.9 & 62.5 & 65.8 & 68.3 & 68.5 & 48.2 & 60.1 & 60.1 & 55.1 & 55.3 & 69.7 & 49.7 & 49.9 & 68.9 \\
Reg3D-AD & 59.7 & 70.6 & 48.6 & 69.5 & 52.5 & 67.1 & 52.5 & 49.0 & 34.8 & 66.3 & 59.3 & 61.0 & 75.2 & 69.3 & 72.5 & 64.3 & 46.7 & 51.0 & 53.8 & 34.3 \\
IMRNet & 67.1 & 66.0 & 55.2 & 70.0 & 64.0 & 68.1 & 70.2 & 68.5 & 59.9 & 67.6 & 71.0 & 58.0 & 77.1 & 73.7 & 77.5 & 65.2 & 65.2 & 64.3 & 75.7 & 54.8 \\
R3D-AD & 83.3 & 72.0 & 73.3 & 73.7 & 78.1 & 81.9 & 77.8 & 74.1 & 76.7 & 74.4 & 65.6 & 68.3 & 75.6 & 82.2 & 73.0 & 68.1 & 67.0 & 77.6 & 75.7 & 89.0 \\
PO3AD & \topone{100.0} & 83.3 & 90.0 & 93.3 & 92.6 & 92.2 & \toptwo{82.9} & 83.3 & 88.1 & \topone{98.1} & 84.9 & 85.3 & 78.7 & 87.7 & 85.9 & 79.2 & 67.0 & 87.1 & 83.3 & \toptwo{99.5} \\
Reg2Inv & 90.0 & \topone{100.0} & \topone{100.0} & \topone{100.0} & \topone{100.0} & \topone{100.0} & 80.7 & 65.6 & 58.5 & 85.2 & 81.8 & 81.3 & \toptwo{90.2} & 65.9 & \toptwo{86.3} & 68.1 & 90.2 & 73.3 & \topone{93.3} & \topone{100.0} \\
AF3AD (Ours) (Seed 1) & \topone{100.0} & \toptwo{94.2} & \topone{100.0} & 99.6 & 96.8 & \toptwo{99.6} & 82.5 & 95.9 & \topone{99.6} & 92.5 & \topone{91.5} & \topone{93.9} & 83.4 & \topone{97.7} & 81.7 & \toptwo{90.8} & \toptwo{92.6} & \topone{100.0} & 85.2 & \topone{100.0} \\
AF3AD (Ours) (Seed 2) & \topone{100.0} & 88.1 & \topone{100.0} & 99.3 & 98.1 & \toptwo{99.6} & \topone{86.1} & \toptwo{96.6} & \toptwo{95.9} & \toptwo{96.7} & 88.1 & 91.1 & 83.8 & \toptwo{95.9} & 77.5 & 86.7 & 86.7 & \topone{100.0} & 84.7 & 95.2 \\
AF3AD (Ours) (Seed 3) & \toptwo{99.0} & 86.7 & \toptwo{93.8} & \toptwo{99.7} & \toptwo{98.7} & 97.8 & 82.2 & \topone{97.4} & 93.7 & 92.2 & \toptwo{90.5} & \toptwo{92.9} & \topone{98.5} & 85.7 & \topone{94.7} & \topone{99.3} & \topone{96.5} & \toptwo{95.7} & \toptwo{92.4} & 97.6 \\
\midrule
Method & Headset0 & Headset1 & Helmet0 & Helmet1 & Helmet2 & Helmet3 & Jar0 & Phone & Shelf0 & Tap0 & Tap1 & Vase0 & Vase1 & Vase2 & Vase3 & Vase4 & Vase5 & Vase7 & Vase8 & Vase9 \\ \midrule
BTF(RAW) & 37.8 & 51.5 & 55.3 & 34.9 & 60.2 & 52.6 & 42.0 & 56.3 & 16.4 & 52.5 & 57.3 & 53.1 & 54.9 & 41.0 & 71.7 & 42.5 & 58.5 & 44.8 & 42.4 & 56.4 \\
BTF(FPFH) & 52.0 & 49.0 & 57.1 & 71.9 & 54.2 & 44.4 & 42.4 & 67.1 & 60.9 & 56.0 & 54.6 & 34.2 & 21.9 & 54.6 & 69.9 & 51.0 & 40.9 & 51.8 & 66.8 & 26.8 \\
M3DM & 57.7 & 61.7 & 52.6 & 42.7 & 62.3 & 37.4 & 44.1 & 35.7 & 56.4 & 75.4 & 73.9 & 42.3 & 42.7 & 73.7 & 43.9 & 47.6 & 31.7 & 65.7 & 66.3 & 66.3 \\
PatchCore(FPFH) & 58.3 & 63.7 & 54.6 & 48.4 & 42.5 & 40.4 & 47.2 & 38.8 & 49.4 & 75.3 & 76.6 & 45.5 & 42.3 & 72.1 & 44.9 & 50.6 & 41.7 & 69.3 & 66.2 & 66.0 \\
PatchCore(PMAE) & 59.1 & 62.7 & 55.6 & 55.2 & 44.7 & 42.4 & 48.3 & 48.8 & 52.3 & 45.8 & 53.8 & 44.7 & 55.2 & 74.1 & 46.0 & 51.6 & 57.9 & 65.0 & 66.3 & 62.9 \\
CPMF & 64.3 & 45.8 & 55.5 & 58.9 & 46.2 & 52.0 & 61.0 & 50.9 & 68.5 & 35.9 & 69.7 & 45.1 & 34.5 & 58.2 & 58.2 & 51.4 & 61.8 & 39.7 & 52.9 & 60.9 \\
Reg3D-AD & 53.7 & 61.0 & 60.0 & 38.1 & 61.4 & 36.7 & 59.2 & 41.4 & 68.8 & 67.6 & 64.1 & 53.3 & 70.2 & 60.5 & 65.0 & 50.0 & 52.0 & 46.2 & 62.0 & 59.4 \\
IMRNet & 72.0 & 67.6 & 59.7 & 60.0 & 64.1 & 57.3 & 78.0 & 75.5 & 60.3 & 67.6 & 69.6 & 53.3 & 75.7 & 61.4 & 70.0 & 52.4 & 67.6 & 63.5 & 63.0 & 59.4 \\
R3D-AD & 73.8 & 79.5 & 75.7 & 72.0 & 63.3 & 70.7 & 83.8 & 76.2 & 69.6 & 73.6 & \topone{90.0} & 78.8 & 72.9 & 75.2 & 74.2 & 63.0 & 75.7 & 77.1 & 72.1 & 71.8 \\
PO3AD & 80.8 & 92.3 & 76.2 & 96.1 & 86.9 & 75.4 & 86.6 & 77.6 & 57.3 & 74.5 & 68.1 & 85.8 & 74.2 & \toptwo{95.2} & 82.1 & 67.5 & 85.2 & 96.6 & 73.9 & 83.0 \\
Reg2Inv & \topone{100.0} & 84.3 & \toptwo{81.7} & \topone{98.6} & 87.5 & \toptwo{87.6} & \topone{100.0} & \topone{100.0} & 57.7 & \topone{94.8} & 80.4 & \topone{99.6} & 60.5 & \topone{100.0} & 84.5 & 81.8 & \topone{100.0} & 64.3 & 81.8 & \toptwo{87.3} \\
AF3AD (Ours) (Seed 1) & \topone{100.0} & \topone{100.0} & 75.9 & 94.2 & 84.3 & 76.0 & \toptwo{99.5} & 95.7 & 61.1 & 83.0 & 82.5 & \toptwo{93.3} & \toptwo{81.4} & \topone{100.0} & \toptwo{92.7} & \toptwo{86.6} & \toptwo{92.3} & \toptwo{99.5} & \topone{88.4} & \topone{87.5} \\
AF3AD (Ours) (Seed 2) & \topone{100.0} & 89.0 & 79.7 & 57.1 & \toptwo{92.8} & 73.9 & 85.7 & 97.6 & \toptwo{72.1} & \toptwo{85.4} & 84.8 & \toptwo{93.3} & 80.5 & \topone{100.0} & 90.9 & 84.8 & 91.4 & \topone{100.0} & \toptwo{87.0} & 83.9 \\
AF3AD (Ours) (Seed 3) & \toptwo{96.8} & \toptwo{98.9} & \topone{98.8} & \toptwo{97.6} & \topone{96.2} & \topone{98.3} & 97.6 & \toptwo{99.0} & \topone{83.3} & 80.0 & \toptwo{88.9} & \toptwo{93.3} & \topone{96.4} & 95.0 & \topone{94.0} & \topone{93.8} & 86.1 & 96.4 & 85.4 & 86.3 \\
\bottomrule
\end{tabular}%
}
\end{table*}

\begin{table*}[htbp]
    \centering
    \caption{Comparison of P-AUROC results (\%) at the point level. Bold: Best, Underline: Second-best.}
    \label{tab:full_ANS_pauroc_results}
    \footnotesize
    \setlength{\tabcolsep}{1.5pt}
    \renewcommand{\arraystretch}{1.12}
    \resizebox{\textwidth}{!}{%
    \begin{tabular}{l cccccccccccccccccccc}
    \toprule
    \multicolumn{21}{c}{P-AUROC($\uparrow$)} \\ \midrule
    Method & Ashtray0 & Bag0 & Bottle0 & Bottle1 & Bottle3 & Bowl0 & Bowl1 & Bowl2 & Bowl3 & Bowl4 & Bowl5 & Bucket0 & Bucket1 & Cap0 & Cap3 & Cap4 & Cap5 & Cup0 & Cup1 & Eraser0 \\ \midrule
    BTF(RAW) & 51.2 & 43.0 & 55.1 & 49.1 & 72.0 & 52.4 & 46.4 & 42.6 & 68.5 & 56.3 & 51.7 & 61.7 & 68.6 & 52.4 & 68.7 & 46.9 & 37.3 & 63.2 & 56.1 & 63.7 \\
    BTF(FPFH) & 62.4 & 74.6 & 64.1 & 54.9 & 62.2 & 71.0 & 76.8 & 51.8 & 59.0 & 67.9 & 69.9 & 40.1 & 63.3 & 73.0 & 65.8 & 52.4 & 58.6 & 79.0 & 61.9 & 71.9 \\
    M3DM & 57.7 & 63.7 & 66.3 & 63.7 & 53.2 & 65.8 & 66.3 & 69.4 & 65.7 & 62.4 & 48.9 & 69.8 & 69.9 & 53.1 & 60.5 & 71.8 & 65.5 & 71.5 & 55.6 & 71.0 \\
    PatchCore(FPFH) & 59.7 & 57.4 & 65.4 & 68.7 & 51.2 & 52.4 & 53.1 & 62.5 & 32.7 & 72.0 & 35.8 & 45.9 & 57.1 & 47.2 & 65.3 & 59.5 & 79.5 & 65.5 & 59.6 & 81.0 \\
    PatchCore(PMAE) & 49.5 & 67.4 & 55.3 & 60.6 & 65.3 & 52.7 & 52.4 & 51.5 & 58.1 & 50.1 & 56.2 & 58.6 & 57.4 & 54.4 & 48.8 & 72.5 & 54.5 & 51.0 & 85.6 & 37.8 \\
    CPMF & 61.5 & 65.5 & 52.1 & 57.1 & 43.5 & 74.5 & 48.8 & 63.5 & 64.1 & 68.3 & 68.4 & 48.6 & 60.1 & 60.1 & 55.1 & 55.3 & 55.1 & 49.7 & 50.9 & 68.9 \\
    Reg3D-AD & 69.8 & 71.5 & 88.6 & 69.6 & 52.5 & 77.5 & 61.5 & 59.3 & 65.4 & 80.0 & 69.1 & 61.9 & 75.2 & 63.2 & 71.8 & 81.5 & 46.7 & 68.5 & 69.8 & 75.5 \\
    IMRNet & 67.1 & 66.8 & 55.6 & 70.2 & 64.1 & 78.1 & 70.5 & 68.4 & 59.9 & 57.6 & 71.5 & 58.5 & 77.4 & 71.5 & 70.6 & 75.3 & 74.2 & 64.3 & 68.8 & 54.8 \\
    ISMP & 60.3 & 74.7 & 77.0 & 56.8 & 77.5 & 85.1 & 54.6 & 73.6 & 77.3 & 74.0 & 53.4 & 52.4 & 67.2 & 86.5 & 73.4 & 75.3 & 67.8 & 86.9 & 60.0 & 70.6 \\
    PO3AD & \topone{96.2} & \toptwo{94.9} & 91.2 & 84.4 & \toptwo{88.0} & 97.8 & \topone{91.4} & 91.8 & 93.5 & \toptwo{96.7} & \toptwo{94.1} & \toptwo{75.5} & 89.9 & 95.7 & 94.8 & 94.0 & 86.4 & 90.9 & \topone{93.2} & 97.4 \\
    Reg2Inv & 78.5 & \topone{99.1} & \topone{99.5} & 84.9 & 81.7 & \topone{98.3} & 82.8 & 82.2 & 76.1 & 78.8 & 82.4 & 61.0 & 85.5 & 86.1 & 94.5 & 86.4 & 97.0 & 79.8 & 88.1 & \topone{98.0} \\
    AF3AD (Ours) (Seed 1) & \toptwo{95.0} & 94.0 & \toptwo{99.0} & \toptwo{93.8} & \topone{92.6} & \toptwo{97.9} & 87.8 & \toptwo{96.2} & \topone{98.3} & 96.5 & \topone{97.4} & 70.6 & \toptwo{95.6} & \topone{98.6} & \topone{98.9} & \topone{99.0} & \topone{98.0} & \topone{95.1} & \toptwo{91.9} & 95.4 \\
    AF3AD (Ours) (Seed 2) & 79.4 & 92.1 & 98.6 & \topone{95.0} & 85.7 & 95.2 & \toptwo{91.1} & \topone{97.9} & \toptwo{97.7} & \topone{96.9} & 91.8 & 65.5 & 93.0 & \toptwo{98.4} & \toptwo{97.1} & \toptwo{97.6} & \toptwo{97.9} & \topone{95.1} & 88.4 & 90.8 \\
    AF3AD (Ours) (Seed 3) & 76.8 & 93.0 & 98.2 & 93.2 & 87.7 & 89.3 & 88.7 & \topone{97.9} & 95.0 & 94.3 & 92.7 & \topone{91.7} & \topone{97.7} & 90.0 & 93.3 & 97.4 & 91.9 & \toptwo{91.3} & 84.4 & \toptwo{97.6} \\
    \midrule
    Method & Headset0 & Headset1 & Helmet0 & Helmet1 & Helmet2 & Helmet3 & Jar0 & Phone & Shelf0 & Tap0 & Tap1 & Vase0 & Vase1 & Vase2 & Vase3 & Vase4 & Vase5 & Vase7 & Vase8 & Vase9 \\ \midrule
    BTF(RAW) & 57.8 & 47.5 & 50.4 & 44.9 & 60.5 & 70.0 & 42.3 & 58.3 & 46.4 & 52.7 & 56.4 & 61.8 & 54.9 & 40.3 & 60.2 & 61.3 & 58.5 & 57.8 & 55.0 & 56.4 \\
    BTF(FPFH) & 62.0 & 59.1 & 57.5 & 74.9 & 64.3 & 72.4 & 42.7 & 67.5 & 61.9 & 56.8 & 59.6 & 64.2 & 61.9 & 64.6 & 69.9 & 71.0 & 42.9 & 54.0 & 66.2 & 56.8 \\
    M3DM & 58.1 & 58.5 & 59.9 & 42.7 & 62.3 & 65.5 & 54.1 & 35.8 & 55.4 & 65.4 & 71.2 & 60.8 & 60.2 & 73.7 & 65.8 & 65.5 & 64.2 & 51.7 & 55.1 & 66.3 \\
    PatchCore(FPFH) & 58.3 & 46.4 & 54.8 & 48.9 & 45.5 & 73.7 & 47.8 & 48.8 & 61.3 & 73.3 & 76.8 & 65.5 & 45.3 & 72.1 & 43.0 & 50.5 & 44.7 & 69.3 & 57.5 & 66.3 \\
    PatchCore(PMAE) & 57.5 & 42.3 & 58.0 & 56.2 & 65.1 & 61.5 & 48.7 & 88.6 & 54.3 & 85.8 & 54.1 & 67.7 & 55.1 & 74.2 & 46.5 & 52.3 & 57.2 & 65.1 & 36.4 & 42.3 \\
    CPMF & 69.9 & 45.8 & 55.5 & 54.2 & 51.5 & 52.0 & 61.1 & 54.5 & \toptwo{78.3} & 45.8 & 65.7 & 45.8 & 48.6 & 58.2 & 58.2 & 51.4 & 65.1 & 50.4 & 52.9 & 54.5 \\
    Reg3D-AD & 58.0 & 62.6 & 60.0 & 62.4 & 82.5 & 62.0 & 59.9 & 59.9 & 68.8 & 58.9 & 74.1 & 54.8 & 60.2 & 40.5 & 51.1 & 75.5 & 62.4 & 88.1 & 81.1 & 69.4 \\
    IMRNet & 70.5 & 47.6 & 59.8 & 60.4 & 64.4 & 66.3 & 76.5 & 74.2 & 60.5 & 68.1 & 69.9 & 53.5 & 68.5 & 61.4 & 40.1 & 52.4 & 68.2 & 59.3 & 63.5 & 69.1 \\
    ISMP & 58.0 & 70.2 & 68.3 & 62.2 & 84.4 & 72.2 & 82.3 & 66.1 & 68.7 & 52.2 & 55.2 & 66.1 & 84.3 & 73.3 & 76.2 & 54.5 & 47.2 & 70.1 & 85.1 & 61.5 \\
    PO3AD & 82.3 & 90.7 & 87.8 & \toptwo{94.8} & 93.2 & 84.6 & 87.1 & 81.0 & 66.3 & 78.3 & 69.2 & 95.5 & \toptwo{88.2} & 97.8 & 88.4 & 90.2 & \toptwo{93.7} & 98.2 & 95.0 & 95.2 \\
    Reg2Inv & 94.6 & \topone{97.0} & \toptwo{92.5} & 90.6 & 89.1 & \toptwo{95.6} & \toptwo{98.2} & \toptwo{99.2} & 63.2 & \topone{91.8} & 86.9 & 98.0 & 70.5 & \topone{99.7} & 84.4 & 92.7 & 87.9 & 86.3 & 93.4 & 97.1 \\
    AF3AD (Ours) (Seed 1) & \toptwo{97.2} & \toptwo{95.2} & 91.8 & 76.8 & 97.1 & 81.1 & \topone{99.1} & 92.8 & 69.6 & 82.7 & \toptwo{91.5} & \toptwo{99.1} & 87.2 & \toptwo{99.2} & \topone{97.3} & \topone{97.7} & \topone{95.3} & \topone{99.2} & 96.6 & 97.4 \\
    AF3AD (Ours) (Seed 2) & 95.4 & 81.9 & 91.7 & 59.6 & \toptwo{97.2} & 73.4 & 85.7 & 90.3 & 76.0 & \toptwo{89.9} & 87.7 & \topone{99.2} & 88.0 & 97.5 & 94.5 & \toptwo{96.9} & 83.0 & \toptwo{99.0} & \toptwo{97.4} & \topone{97.6} \\
    AF3AD (Ours) (Seed 3) & \topone{97.3} & 92.8 & \topone{98.9} & \topone{97.7} & \topone{97.8} & \topone{99.0} & 93.9 & \topone{99.7} & \topone{95.2} & 85.7 & \topone{92.8} & 98.7 & \topone{92.8} & 97.0 & \toptwo{96.5} & 94.9 & 80.1 & 96.9 & \topone{97.6} & \toptwo{97.5} \\
    \bottomrule
    \end{tabular}%
    }
    \end{table*}
\begin{table*}[tbp]
\centering
\caption{Comparison of O-AUROC results (\%) on Real3D-AD. Bold: Best, Underline: Second-best.}
\label{tab:full_Real3d_oauroc_results}
\footnotesize
\setlength{\tabcolsep}{1.5pt}
\renewcommand{\arraystretch}{1.12}
\resizebox{\textwidth}{!}{%
\begin{tabular}{l cccccccccccc}
\toprule
\multicolumn{13}{c}{O-AUROC($\uparrow$)} \\ \midrule
Method & Airplane & Car & Candy & Chicken & Diamond & Duck & Fish & Gemstone & Seahorse & Shell & Starfish & Toffees \\ \midrule
BTF(RAW) & 73.0 & 64.7 & 53.9 & 78.9 & 70.7 & 69.1 & 60.2 & 68.6 & 59.6 & 39.6 & 53.0 & 70.3 \\
BTF(FPFH) & 52.0 & 56.0 & 63.0 & 43.2 & 54.5 & 78.4 & 54.9 & 64.8 & 77.9 & 75.4 & 57.5 & 46.2 \\
M3DM & 43.4 & 54.1 & 55.2 & 68.3 & 60.2 & 43.3 & 54.0 & 64.4 & 49.5 & 69.4 & 55.1 & 45.0 \\
PatchCore(FPFH) & \toptwo{88.2} & 59.0 & 54.1 & 83.7 & 57.4 & 54.6 & 67.5 & 37.0 & 50.5 & 58.9 & 44.1 & 56.5 \\
PatchCore(PMAE) & 72.6 & 49.8 & 66.3 & 82.7 & 78.3 & 48.9 & 63.0 & 37.4 & 53.9 & 50.1 & 51.9 & 58.5 \\
CPMF & 70.1 & 55.1 & 55.2 & 50.4 & 52.3 & 58.2 & 55.8 & 58.9 & 72.9 & 65.3 & 70.0 & 39.0 \\
Reg3D-AD & 71.6 & 69.7 & 68.5 & \toptwo{85.2} & 90.0 & 58.4 & 91.5 & 41.7 & 76.2 & 58.3 & 50.6 & 82.7 \\
IMRNet & 76.2 & 71.1 & 75.5 & 78.0 & 90.5 & 51.7 & 88.0 & 67.4 & 60.4 & 66.5 & 67.4 & 77.4 \\
ISMP & 85.8 & 73.1 & 85.2 & 71.4 & 94.8 & 71.2 & 94.5 & 46.8 & 72.9 & 62.3 & 66.0 & 84.2 \\
Group3AD & 74.4 & 72.8 & 84.7 & 78.6 & 93.2 & 67.9 & \topone{97.6} & 53.9 & \toptwo{84.1} & 58.5 & 56.2 & 79.6 \\
R3D-AD & 77.2 & 69.3 & 71.3 & 71.4 & 68.5 & \topone{90.9} & 69.2 & 66.5 & 72.0 & \toptwo{84.0} & 70.1 & 70.3 \\
PO3AD & 80.4 & 65.4 & 78.5 & 68.6 & 80.1 & 82.0 & 85.9 & 69.3 & 75.6 & 80.0 & 75.8 & 77.1 \\
Reg2Inv & 81.8 & 75.8 & \topone{100.0} & \topone{94.4} & \topone{100.0} & 75.0 & 67.2 & 73.5 & 53.2 & 69.2 & 84.1 & 62.6 \\
AF3AD (Ours) (Seed 1) & \topone{90.2} & \toptwo{76.6} & 91.2 & 78.9 & \toptwo{99.3} & \toptwo{87.3} & 96.3 & \toptwo{80.0} & 79.5 & \topone{85.2} & 81.2 & 90.0 \\
AF3AD (Ours) (Seed 2) & 87.0 & 70.5 & \toptwo{92.2} & 72.3 & \topone{100.0} & 66.0 & 95.0 & \topone{83.0} & \topone{86.3} & 76.0 & \toptwo{84.6} & \toptwo{93.1} \\
AF3AD (Ours) (Seed 3) & 71.9 & \topone{90.2} & 90.8 & 77.2 & \topone{100.0} & 76.0 & \toptwo{96.7} & 79.6 & 81.7 & 81.2 & \topone{85.3} & \topone{93.8} \\
\bottomrule
\end{tabular}%
}
\end{table*}

\begin{table*}[htbp]
\centering
\caption{Comparison of P-AUROC results (\%) on Real3D-AD. Bold: Best, Underline: Second-best.}
\label{tab:full_Real3d_pauroc_results}
\footnotesize
\setlength{\tabcolsep}{1.5pt}
\renewcommand{\arraystretch}{1.12}
\resizebox{\textwidth}{!}{%
\begin{tabular}{l cccccccccccc}
\toprule
\multicolumn{13}{c}{P-AUROC($\uparrow$)} \\ \midrule
Method & Airplane & Car & Candy & Chicken & Diamond & Duck & Fish & Gemstone & Seahorse & Shell & Starfish & Toffees \\ \midrule
BTF(RAW) & 56.4 & 64.7 & 73.5 & 60.9 & 56.3 & 60.1 & 51.4 & 59.7 & 52.0 & 48.9 & 39.2 & 62.3 \\
BTF(FPFH) & 73.8 & 70.8 & 86.4 & 73.5 & 88.2 & 87.5 & 70.9 & 89.1 & 51.2 & 57.1 & 50.1 & 81.5 \\
M3DM & 54.7 & 60.2 & 67.9 & 67.8 & 60.8 & 66.7 & 60.6 & 67.4 & 56.0 & 73.8 & 53.2 & 68.2 \\
PatchCore(FPFH) & 56.2 & 75.4 & 78.0 & 42.9 & 82.8 & 26.4 & 82.9 & 91.0 & 73.9 & 73.9 & 60.6 & 74.7 \\
PatchCore(PMAE) & 56.9 & 60.9 & 62.7 & 72.9 & 71.8 & 52.8 & 71.7 & 44.4 & 63.3 & 70.9 & 58.0 & 58.0 \\
CPMF & 61.8 & \toptwo{83.6} & 73.4 & 55.9 & 75.3 & 71.9 & \topone{98.8} & 44.9 & \topone{96.2} & 72.5 & 80.0 & \topone{95.9} \\
Group3AD & 63.6 & 74.5 & 73.8 & 75.9 & 86.2 & 63.1 & 83.6 & 56.4 & 82.7 & 79.8 & 62.5 & 80.3 \\
Reg3D-AD & 63.1 & 71.8 & 72.4 & 67.6 & 83.5 & 50.3 & 82.6 & 54.5 & 81.7 & 81.1 & 61.7 & 75.9 \\
ISMP & 75.3 & \toptwo{83.6} & 90.7 & \toptwo{79.8} & 92.6 & 87.6 & 88.6 & 85.7 & 81.3 & 83.9 & 64.1 & 89.5 \\
Reg2Inv & \topone{92.3} & \topone{94.4} & \topone{96.9} & \topone{91.0} & 97.9 & \topone{93.7} & 84.6 & 90.7 & 64.5 & \topone{90.6} & \topone{84.0} & 73.7 \\
AF3AD (Ours) (Seed 1) & \toptwo{81.3} & 76.4 & 91.1 & 78.8 & \toptwo{98.4} & 79.9 & 93.3 & 88.6 & 70.4 & 82.4 & 79.3 & \toptwo{95.2} \\
AF3AD (Ours) (Seed 2) & 80.6 & 76.6 & \toptwo{94.7} & 77.3 & 98.0 & \toptwo{88.3} & \toptwo{98.0} & \topone{93.4} & \toptwo{88.9} & \toptwo{86.6} & \toptwo{81.6} & 94.9 \\
AF3AD (Ours) (Seed 3) & 72.9 & 74.6 & \toptwo{94.7} & 70.1 & \topone{98.7} & 80.0 & 95.8 & \toptwo{92.8} & 86.4 & 85.3 & 79.1 & 94.8 \\
\bottomrule
\end{tabular}%
}
\end{table*}

\end{document}